\documentclass[review]{elsarticle}

\usepackage{graphicx,epstopdf,times,amsmath,amssymb}
\DeclareGraphicsExtensions{.pdf,.eps,.png,.jpg,.bmp,.mps}

\begin{document}

\begin{frontmatter}

\title{Dimension Reduction with Non-degrading Generalization}

\author{Pitoyo Hartono}
\address{School of Engineering, Chukyo University \\ Yagoto-honmachi Showa-ku Nagoya, Japan}
\ead{hartonot@ieee.com}

\begin{abstract}
Visualizing high dimensional data by projecting them into two or three dimensional space is one of the most effective ways to intuitively understand the data's underlying characteristics, for example their class neighborhood structure. While data visualization in low dimensional space can be efficient for revealing the data's underlying characteristics, classifying a new sample in the reduced-dimensional space is not always beneficial because of the loss of information in expressing the data. It is possible to classify the data in the high dimensional space, while visualizing them in the low dimensional space, but in this case, the visualization is often meaningless because it fails to illustrate the underlying characteristics that are crucial for the classification process.

In this paper, the performance-preserving property of the previously proposed Restricted Radial Basis Function Network in reducing the dimension of labeled data is explained. Here, it is argued through empirical experiments that the internal representation of the Restricted Radial Basis Function Network, which during the supervised learning process organizes a visualizable two dimensional map, does not only preserve the topographical structure of high dimensional data but also captures their class neighborhood structures that are important for classifying them. Hence, unlike many of the existing dimension reduction methods, the Restricted Radial Basis Function Network offers two dimensional visualization that is strongly correlated with the classification process.
\end{abstract}


\end{frontmatter}


\section{Introduction}
In the era of big data, visualization is one of the powerful methods for intuitively discovering the underlying structure of complex data. Since meaningfull data are likely to be multidimensional, visualizing them requires some means of dimension reductions. One of the most traditional dimension reductions algorithms is the Principal Component Analysis (PCA) \cite{Jolliffe}. In PCA, the original coordinate axes of the data are rotated, so that the new axises, the so called Principal Components (PCs), stretch along the distribution of the data in descending order. The data can then be visualized by using the first two or three PCs as new axises. PCA is a method for linearly composing new axises from the original ones by considering the data's distribution in an unsupervised way, i.e. the class labels of the data do not have any role in deciding the PCs. On the other hand, Linear Discriminant Analysis (LDA) \cite{Fisher} generates a new metric distance which maximizes distances between the centroids of subset of data belonging to different classes while at the same time minimizes the dispersions of the subsets of data belonging to a same class. Generally, for $N$-classes data, the maximum rank for the transformation matrix is $N-1$, so by limiting the rank of transformation matrix to 1 to 3, the dimension of the data can be reduced and visualized.  In the general sense of classification, the LDA-reduced space is more descriptive than that of PCA, in that it offers the visualization of the underlying structure of the data in the context of their categories, whereas in PCA, the visualization is detached from the actual class structure of the data. Naturally, applying LDA to categorical data and then classifying new data points using the generated distance metric often yields significantly better classification rate compared to when classifying them in the eigen space produced by PCA, although some exceptions were argued \cite{martinez}. Regardless of the reduction methods, when the dimension of the data is reduced to 2 or 3 for the purpose of the visualization, the classification performance in the reduced dimension can catastrophically degenerate. It means that the visualization in the reduced dimension does not reflect the actual class structure of the data in their original dimension.

This paper tries to argue that the previously proposed Restricted Radial Basis Function Network (rRBF) \cite{Hartono, Trappenberg} offers two dimensional representation of high dimensional categorical data without compromising the classification fidelity. The rRBF is a hierarchical supervised neural network that during its learning process generates a two dimensional internal representation called  Context-Relevant Self-Organizing Maps (CRSOM) that reflects the topographical relation of the given data in the context of their class labels. As this internal representation is two dimensional, it can be readily visualized and is useful in understanding the class structure of the data in the space where the classification takes place. Although the basic characteristics of the rRBF have been previously introduced, the correlation between its visualization and generalization performance is not sufficiently studied and tested against other dimension reduction methods, which will be the primary objective of this paper.

Other than PCA and LDA there are rich collections of dimension reduction methods \cite{vandermaaten}. The traditional ones include the Multidimensional Scalling (MDS) \cite{Kruskal, Sammon}. The objective of MDS and its variants, is to map high dimensional data into a low dimensional space by preserving a criterion of distance in their original high dimension. This criterion can be distance in a well defined metric or a subjective qualitative measure. While it is possible for MDS to reduce the dimension of data while preserving some relations of the data, it does not produce a transformation matrix to map new data into the reduced-dimension space, hence it is not possible to use it for classification.  More recently, many dimensional reduction and metric learnings \cite{Weinberger} have been proposed. For examples, Locally Linear Embedding (LLE)  \cite{Roweis} is a dimension reduction method that preserves the piecewise linearity of the data. It assumes that a data point is the weighted sum of its neighbors in the data's original dimensions, and preserves the weights in the low dimensional space.  In the Stochastic Neighborhood Embedding (SNE) \cite{Hintonroweis} and its variant t-SNE \cite{vandermaatent-sne}, the stochastic relationships of the data, which is the probability that a point is in the neighbor of other points, in the original dimension is preserved in the reduced dimension space. LLE, SNE, and t-SNE are proposed based on very elegant mathematical fondations, but as in the MDS, they do not offer transformation matrix to map new data points into the reduced dimension space. While they offer strong alternatives for PCA in dimension reduction, they cannot visualize the underlying class structure of the data because the class labels are not utilized. Neighborhood Component Analysis (NCA) \cite{Goldberger} is an elegant algorithm to learn a distance metric that maximizes the probability of the data being successfully classified when the Nearest Neighbors classification \cite{Cover} is executed utilizing the learned distance metric. Unlike the LDA, the maximum number of dimensions in NCA is not limited by the number of classes of the data. However, similar to LDA, although it provides a transformation matrix, the successful classification in high dimensional space with the learned distance metric does not guarantee the successful classification in the reduced dimensional space where the data can be visualized. In the occurance of the catastrophic degeneration of the classification performance in the low dimensional space, the visualization using NCA offers no insight for understanding the class structure of the data. In this paper, it is empirically shown that the rRBF does not suffer from this problem.

The output of the hidden layers of hierarchical neural networks can also be used to reduce the dimension of the input. For example, autoencoder composed from deep layers network \cite{Hinton, Hinton2015}, where one of the layers contains two neurons, can be trained and used to produce two dimensional mapping of the high dimensional input. While with this mechanism new data points can be projected into the map, however, similar to PCA, for categorical data the two dimensional map is detached from the class structure of the data. It is obviously possible to train a multilayer classifier where one of the layers contains two neurons that can be used to visualize the high dimensional input. In this case, there will be some correlation between the two dimensional internal representation and the labels of the data. However, due to the complexity of the internal representation, for example the one generated by the iterative executions of Restricted Boltzmann Machine (RBM) \cite{HintonBoltzmann}, the relationship between the original high dimension inputs and their low dimensional representation is unclear. The rRBF offers more comprehensive relation between the high dimensional input and the reduced dimension representations.

The paper is composed as follows. Section 2 is dedicated for explaining the structure and the learning process of the Restricted Radial Basis Function Network. In section 3, experiments where rRBF was compared against PCA, LDA and NCA are explained. For comparing the generalization performance, Nearest Neighbors \cite{Cover} classifications were executed in the reduced dimensions where PCA, LDA and NCA were executed. Conclusions and future works are discussed in the final section.

\section{Restricted RBF with 2-Dimensional Internal Representation}

\begin{figure}[htbp]
  \begin{center}
	\includegraphics[width=8cm,height=8cm]{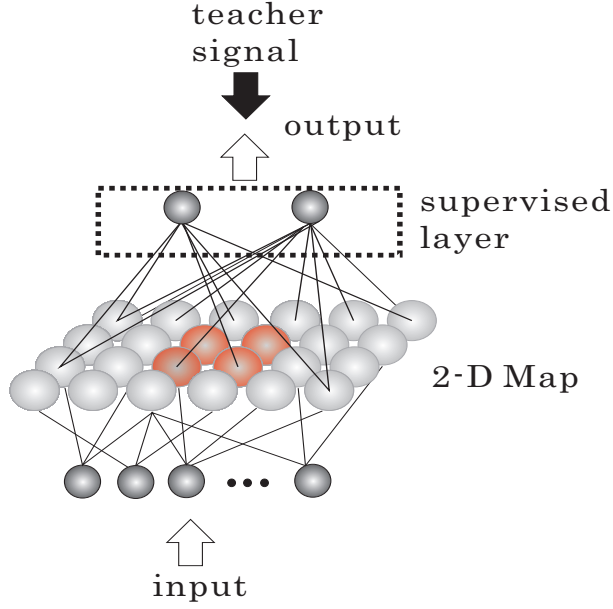}
        \caption{Outline of rRBF \label{fig:rrbf}}
  \end{center}
\end{figure}

Restricted Radadial Basis Function Network (rRBF), shown in in Fig. \ref{fig:rrbf}, is a hierarchical neural network inspired by the conventional Radial Basis Function Networks (RBF) \cite{Pogio090}

The internal layer of the rRBF is a two dimensional grid of neurons, similar to the Self-Organizing Maps (SOM) \cite{Kohonen082}, where the $j$-th neuron is associated with a reference vector, $W_j$ with the same dimensionality as the input. Receiving input $X \in R^d$, at time $t$, the winner, $win$ among the hidden neurons is calculated as follows.

\begin{eqnarray}
win(t) &=& \arg \min_{j} I_j(t) \\
I_j(t) &=& \|X(t)-W_j(t)\| \nonumber
\end{eqnarray}

The output of the $j$-th hidden neuron, $O^{hid}_j$ and the $k$-th output of the rRBF, $O_k(t)$, at time $t$, are then calculated as follows.

\begin{equation}
O^{hid}_j(t) = \sigma(win,j,t) e^{-I_j(t)} \label{eq:outhid}
\end{equation}

In Eq. \ref{eq:outhid} $\sigma(win,j,t)$ is a neighborhood function defined as,

\begin{eqnarray}
 \sigma(win,j,t) &=& e^{-\frac{dist(win,j)}{S(t)}} \label{eq:neighbor} \\
   S(t)            &=& S_{start} ( \frac{S_{end}}{S_{start}})^{\frac{t}{t_{end}}}   \hspace {2mm} (0 \le t \le t_{end}, S_{start} > S_{end}) \nonumber
\end{eqnarray}

where $dist(win,j,t)$ is the distance from the winning neuron to the $j$-th neuron in the two-dimensional grid, while $t$, and $t_{end}$, is the current epoch, and the target epoch when the learning process is terminated. The difference between the rRBF and the conventional RBF is that in the rRBF the output of the hidden neurons are topologically restricted by the winning neuron through the neighborhood function.

\begin{equation}
O_k(t) = f(\sum_j v_{jk} O^{hid}_j - \theta_k) \label{eq:out}
\end{equation}

In Eq. \ref{eq:out}, $v_{jk}$ and $\theta_k$ are the weight connecting the $j$-th hidden neuron and the $k$-th output neuron, and the bias of the output neuron, respectively.

The error function is defined as follows.

\begin{equation}
E(t) = \frac{1}{2} \sum_k (T_k(t) - O_k(t))^2
\end{equation}

Here $T_k$ is the $k$-th component of the teacher signal.

Executing gradient descent as in Backpropogation \cite{Rumelhart1984, Rumelhart1986}, the corrections of the connection weights between the hidden and output layer and the $i$-th reference vector, $W_i$, in the hidden neurons can be calculated as follows.

\begin{eqnarray}
\frac{\partial E(t)}{\partial v_{jk}(t)} &=& \delta^{out}_k(t) O^{hid}_j(t) \\
\frac{\partial E(t)}{\partial \theta_k(t)} &=&  -\delta^{out}_k(t)  \\
 \delta^{out}_k(t)    &=& -(T_k(t) - O_k(t)) O_k(t) (1 - O_k(t)) \nonumber
\end{eqnarray}

\begin{equation}
\frac{\partial E(t)}{\partial W_i(t)} = -(\sum_k \delta^{out}_k(t) v_{ik}) O^{hid}_i(t) (X(t)-W_i(t)) 
\end{equation}

Hence, the reference vector is modified according to Eq. \ref{eq:refmodify}.

\begin{eqnarray}
W_i(t+1) &=& W_i(t) + \eta \delta^{hid}_i \sigma(win,i,t) (X(t)-W_i(t)) \label{eq:refmodify} \\
\delta^{hid}_i(t) &=& e^{-I_i(t)} \sum_k \delta^{out}_k(t) v_{ik}(t) \nonumber
\end{eqnarray}

Although the formula for the modification of the reference vectors in Eq. \ref{eq:refmodify} is similar to that of traditional SOM, the inclusion of the term $\delta^{out}_k(t)$ significantly influences the map formation. In organizing the map, the conventional SOM does not access the data labels, hence the generated map preserves ony the topological similarities of the data while ignoring their class neighborhood structure. In Eq. \ref{eq:refmodify}, $\delta^{out}_k(t)$ is the backpropagated error information from the output layer. Since, this term is influenced by the output of the network and the true label of the input, it includes the categorical information of the data. Without loss of generality, the influence of this term is better to be explained in the case of two-class problems. In this case, two similar inputs with opposing classes will likely generate $\delta^{out}$ with opposing signs. Since in this case the winning hidden neuron is likely to be the same, the opposing sign will result in the reference vector to be modified towards one of the input, while repelled from the other. This dynamics will generate margins between the projections of the two inputs, hence the generated map does not only preserve the topological neighborhood but also the class neighborhood of the data. As the output of this hidden layer is propagated to the output layer, the classification results of the rRBF is a function of the activity in the map, which means that the map visualizes the actual problem space where the decision is being made. This visualization characteristic significantly distinguish rRBF from many visualization methods that are often detached from the actual classification process.

\section{Experiments}

\subsection{Classification in the Reduced Dimension}
 In this preliminary experiment, classification of MNIST hand writing data set (10 classes, 784 features) \cite{LeCun} was performed in the reduced dimension space, where the traditional dimension reduction methods of PCA and LDA were applied. These two methods were chosen because of their contrasting natures in reducing the dimension of the data, in which the former does not access the labels of the data, while the later does.

\begin{figure}[htbp]
  \begin{minipage}{0.5\hsize}
  	\begin{center}
		\includegraphics[width=6.5cm,height=5cm]{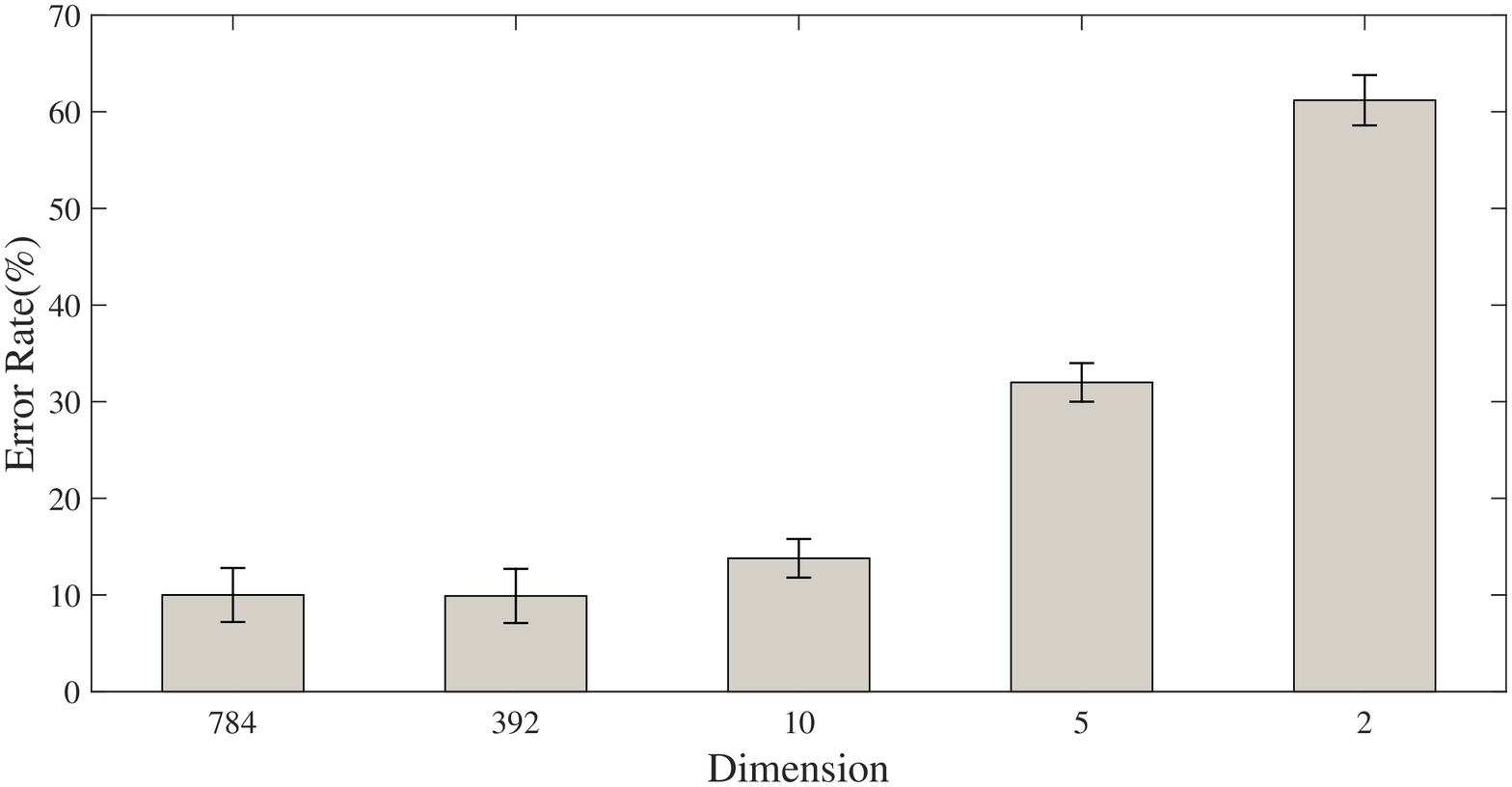}
       		 \caption{Classification MNIST (PCA) \label{fig:mnistpca}}
 	 \end{center}
    \end{minipage}
   \begin{minipage}{0.5\hsize}
  	\begin{center}
		\includegraphics[width=6.5cm,height=5cm]{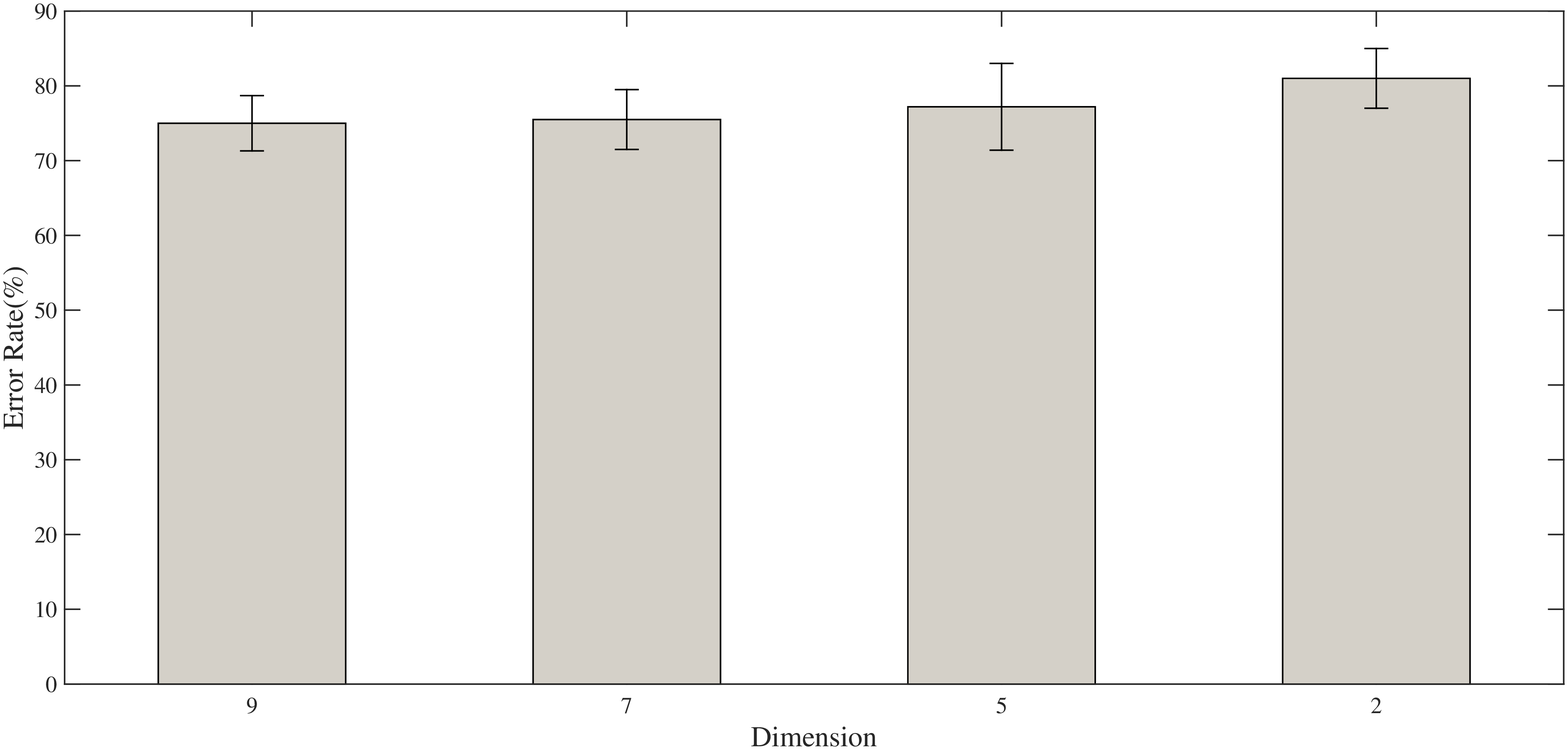}
       		 \caption{Classification MNIST (LDA) \label{fig:mnistlda}}
 	 \end{center}
    \end{minipage}
\end{figure}

Figure \ref{fig:mnistpca} shows the error rate when the nearest neighbors classification was executed on the PCA-reduced dimension against MNIST problem. As the original dimension of this problem is 784, and since PCA is a linear transformation, executing nearest neighbor classification in the 784 PCA-transformed dimension is equivalent with doing so in the original dimension. Figure \ref{fig:mnistpca}  shows that the classification in two dimension is about 6 times worse than the classification in the original dimension. The significant difference in generalization performances in the original and the reduced dimension indicate that the representation in the reduced dimension fails to retain the class structure of the problems, which makes the visualization useless. Figure \ref{fig:mnistlda} shows the result of the nearest neighbors classification in LDA-reduced dimension. As the number of classes for this problem is 10, the maximum allowed number of the reduced dimension is 9. This figure shows that the classification error is high even on the allowed maximum dimensions space. While it is true that the classification of high dimensional data does not have to be performed in the reduced dimension, the visualization of the data in three or two dimensional space does not truely illustrate the characteristics of the data in their original dimension. Hence, in the case of high dimensional with many non-correlated components like MNIST, there is a trade-off between the classification performance and the fidelity of the visualization. 

\begin{figure}[htbp]
  \begin{minipage}{0.5\hsize}
  	\begin{center}
		\includegraphics[width=6.5cm,height=5cm]{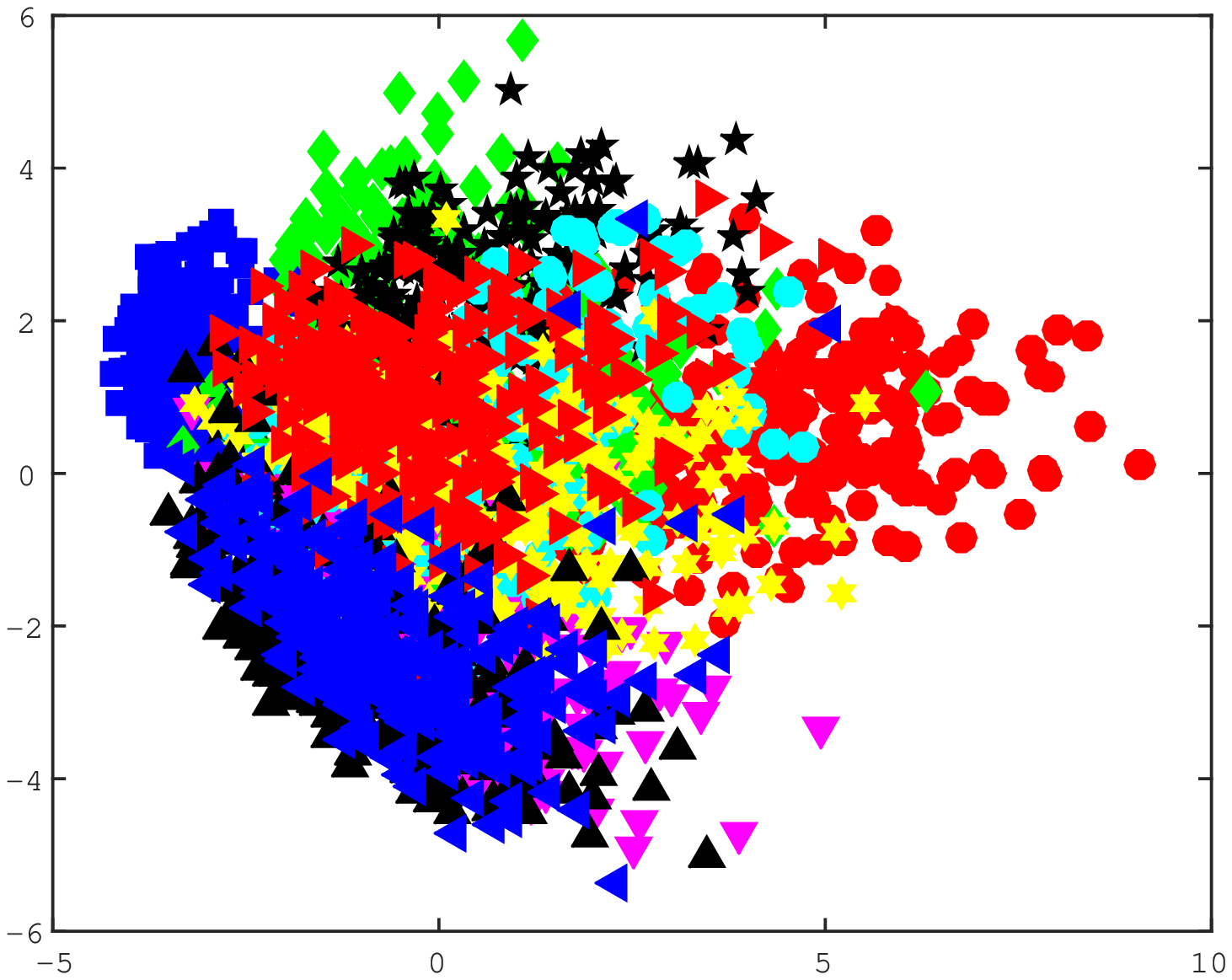}
       		 \caption{PCA: 2-D \label{fig:pcamnist}}
 	 \end{center}
    \end{minipage}
   \begin{minipage}{0.5\hsize}
  	\begin{center}
		\includegraphics[width=6.5cm,height=5cm]{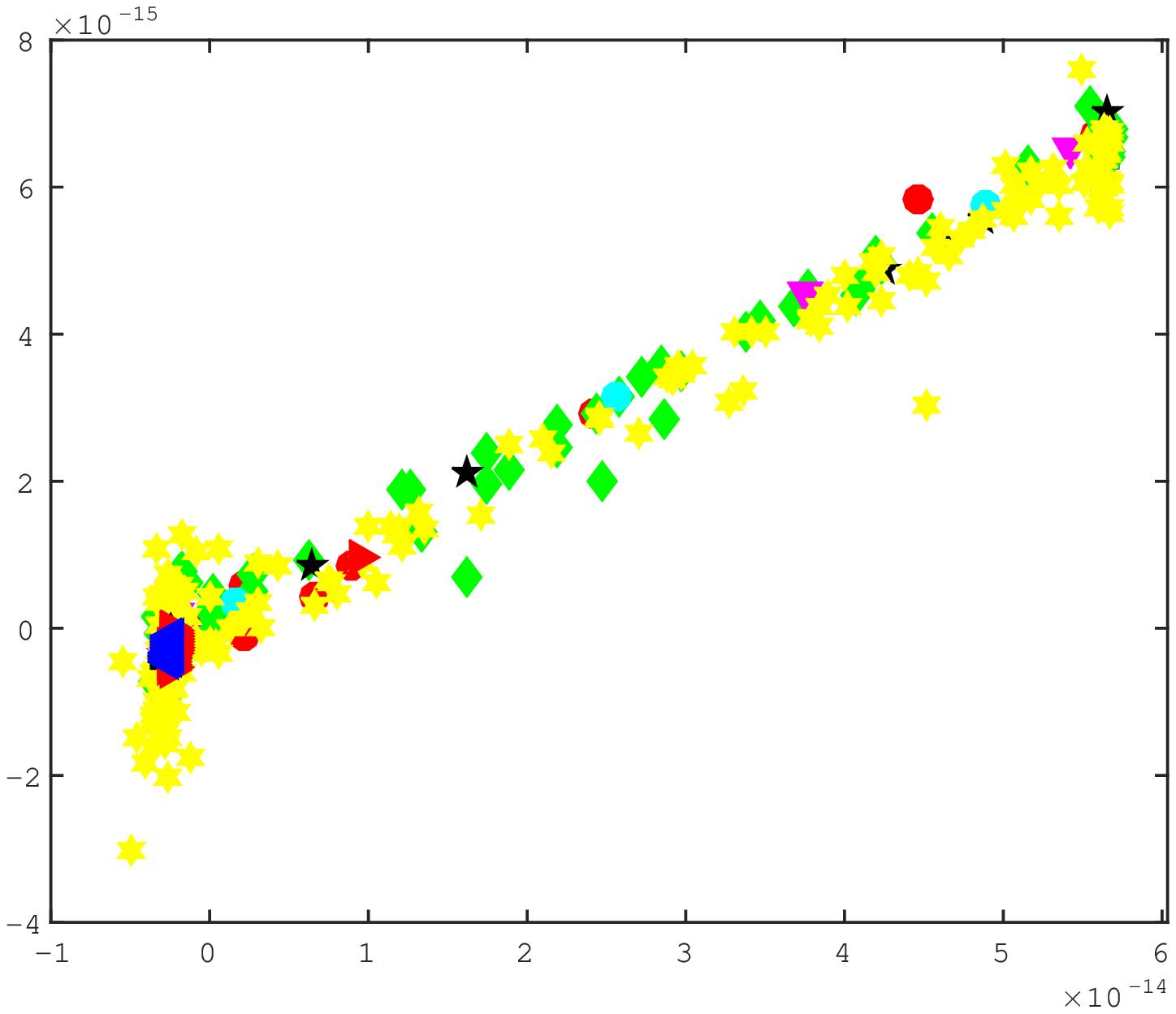}
       		 \caption{LDA: 2-D\label{fig:ldamnist}}
 	 \end{center}
    \end{minipage}
\end{figure}

Figure \ref{fig:pcamnist} shows the distribution of MNIST data in 2 dimension eigen space of PCA, where each class is represented with different color and shape. It is obvious from this figure that in two dimensional eigenspace there are no distinctive clusters of classes, hence consequently nearest neighbor classification produces a large error value, as indicated in Fig. \ref{fig:mnistpca}. Figure \ref{fig:ldamnist} shows the projection of the problem into two dimensional space, where a projection matrix with the rank of 2 was trained using LDA. The two dimensional projection of LDA  did not generate any obvious clusters either. which consequently produced a large classification error as indicated in Fig. \ref{fig:mnistlda}.

This preliminary experiments illustrate the trade-off between visibility of high dimensional data in low dimensional space and their classificability. While MNIST data set are not difficult to classify in their original dimension, as indicated by the low error rate as shown in Fig. \ref{fig:mnistpca}, trying to classify them in two dimensional space produced a poor result. It can be argued that the two dimensional representation through PCA, which does not access the data labels, and LDA, which utilizes the labels, failed to preserve the underlying neighborhood structure of the data. Hence, in this case, the visualization in two dimensional space is not useful in understanding the data. 

\subsection{Comparison Tests}

Here, the performance of the rRBF in visualizing high dimensional data through its internal representation and its generalization performance are empirically compared with those of PCA, LDA and NCA against datasets shown in Table. \ref{tab:dataset}. Table \ref{tab:error} shows the means and standard deviation in the bracket for each methods when 10-fold cross validation tests were executed. Here, rRBF was compared against nearest neighbors classification on visualizable two dimensional space in which the dimension reduction was done using PCA, LDA and NCA, denoted as PCA(2-D), LDA(1,2-D) and NCA (2-D) in the Table \ref{tab:error}. It should be noted that for two classes problems, the LDA projects the data into a one dimensional space. The lowest classification errors in the reduced dimensions are highlighted in bold.

Although the focus of this experiment is to compare the generalization performances of those methods in reduced, and thus visualizable, problem space, as references the classifications in each problem's original dimension are also given. In Table \ref{tab:error}, {\bf NCA} denotes the classification in the original space, where a new distance metric was obtained by NCA is utilized, and {\bf NN} shows the nearest neighbor classification in the original dimensional space.

\begin{table}[htbp]
 \begin{center}
  \caption{Datasets}\label{tab:dataset}
  \begin{tabular}{|c|c|c|c|}
    \hline
    Dataset & Dimension & Classe & Instances \tabularnewline \hline \hline
    Iris & 4 & 3 & 150 \tabularnewline \hline
   Wine & 13 & 3 & 178 \tabularnewline \hline
   Fertility & 9 & 2 & 100 \tabularnewline \hline
   Bupa & 6 & 2 & 345 \tabularnewline \hline
  Thyroid & 5 & 3 & 215 \tabularnewline \hline
  Pima & 8 & 2 & 200 \tabularnewline \hline
  Music & 2508 & 2 & 866 \tabularnewline \hline
  B-cancer & 9 & 2 & 683 \tabularnewline \hline
  Balance & 4 & 3 & 625 \tabularnewline \hline
  Hayesroth & 4 & 3 & 132 \tabularnewline \hline
  (subset of ) MNIST & 784 & 10 & 2499 \tabularnewline \hline
  ISOLET & 617 & 26 & 6238 \tabularnewline \hline
  \end{tabular}
 \end{center}
\end{table}

\begin{table}[htbp]
  \begin{center}
   \caption{Error Rate (\%) (Standard Deviation)} \label{tab:error}
   \begin{tabular}{|c|c|c|c|c||c|c|c|}
      \hline
      Dataset & rRBF & PCA(2-D) & LDA(1,2-D) & NCA (2-D) & {\bf NCA} & {\bf NN}  \tabularnewline \hline \hline  
      Iris & {\bf 2.7 (4.7)} & 3.3 (4.7) & 4.7 (5.5) & 4.7 (5.5) & 5.3 (5.3) & 4.0 (4.7) \tabularnewline \hline
     Wine & {\bf 3.9 (4.6)} & 6.0 (7.0) & 7.8 (4.7) & 7.9 (6.1)  & 2.7 (3.9) & 3.9 (4.6) \tabularnewline \hline
     Fertility & 16 (9.7) & 14.0 (15.1) & {\bf 14.0 (11.7)} & 18 (13.2) & 18.0 (11.4) & 15.0 (12.7) \tabularnewline \hline
     Bupa & {\bf 33.3 (9.3)} & 45.5 (9.6) & 36.3 (7.3) & 44.9 (9.7) & 37.7 (7.8) & 35.1 (8.5) \tabularnewline \hline
    Thyroid & 5.6 (6.0) & 6.1 (5.5) & 6.9 (4.4) & {\bf 4.7 (3.8)} & 6.5 (5.8) & 6.0 (4.4) \tabularnewline \hline
    Pima & {\bf 29.5 (9.6)} & 31.5 (14.0) & 30.5 (10.1) & 35.0 (2.4)  & 31.5 (9.1) & 38.5 (11.1) \tabularnewline \hline
    Music & {\bf 21.0 (4.2)} & 30.3 (5.2) & 45.3 (4.8) & 36.1 (2.4) & 19.9 (3.8) & 37.6 (5.3) \tabularnewline \hline
   B-Cancer & 3.7 (2.4) & {\bf 3.4 (1.7)} & 3.5 (2.1) & 4.1 (2.5) & 5.1 (3.0) & 3.5 (1.9) \tabularnewline \hline
    Balance & 14.2 (4.6) & 46.7 (20.3) & 9.8 (3.7) & {\bf 9.6 (4.9)} & 9.3 (5.4) & 29.4 (5.2) \tabularnewline \hline
    Hayesroth & 35.7 (13.1) & 47.7 (17.8) & {\bf 31.0 (13.0)} & 35.5 (14.9)  & 37.1 (13.5) & 41.8 (18.3) \tabularnewline \hline
   MNIST & {\bf 10.3 (2.0)} & 61.2 (2.6) & 81.0 (4.0) & 50.8 (8.8) & 7.7 (1.3) & 10.0 (2.8) \tabularnewline \hline
  ISOLET & {\bf 19.5 (2.6)} & 74.3 (1.6) & 57.2 (1.7) & 75.6 (1.3) & 2.9  & 17.6 (1.4) \tabularnewline \hline
   \end{tabular}
 \end{center}
\end{table}

Generalization results in Table \ref{tab:error} indicate that for problems with relatively low dimensions, rRBF did not always outperform other methods although its performance is never to far from the best performing methods.The rRBF performed significantly better than other methods for high dimensional problems like MNIST, Music and ISOLET. This is due to the oversimplified representation of those data in two dimensional space by the conventional dimension reduction methods., for example in case of PCA, it is shown by the low cumulative contribution rates of the two highest principal components. The rRBF is not exposed to this drawback, since the internal representation, CRSOM, is not based on the reduced features of the data, but context-oriented alignment of high dimensional data in two dimensional space.

The visualizations of some of the problems are given as follows.

\begin{figure}[htbp]
  \begin{minipage}{0.5\hsize}
  	\begin{center}
		\includegraphics[width=6.5cm,height=5cm]{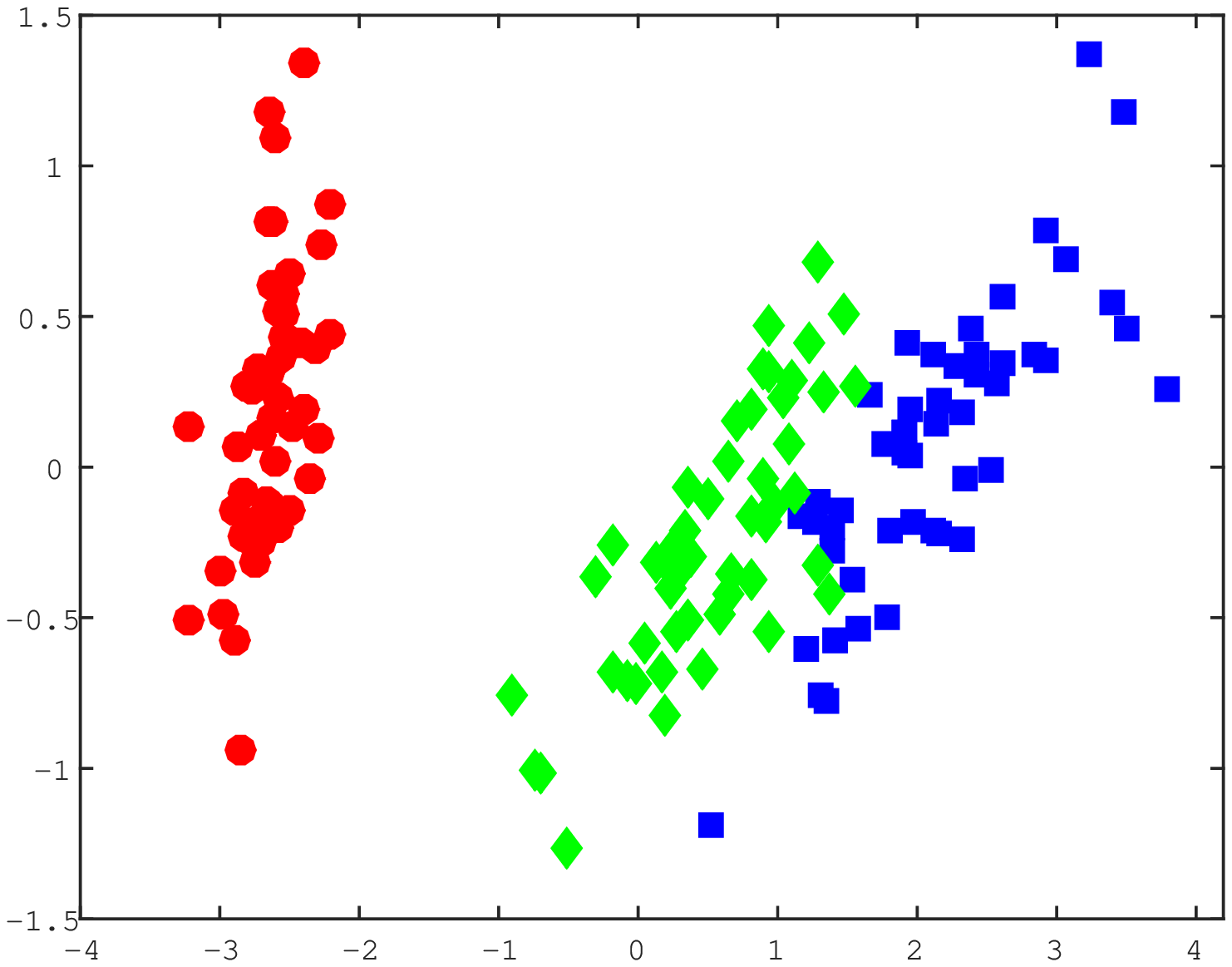}
       		 \caption{PCA (Iris): 2-D\label{fig:pcairis}}
 	 \end{center}
    \end{minipage}
   \begin{minipage}{0.5\hsize}
  	\begin{center}
		\includegraphics[width=6.5cm,height=5cm]{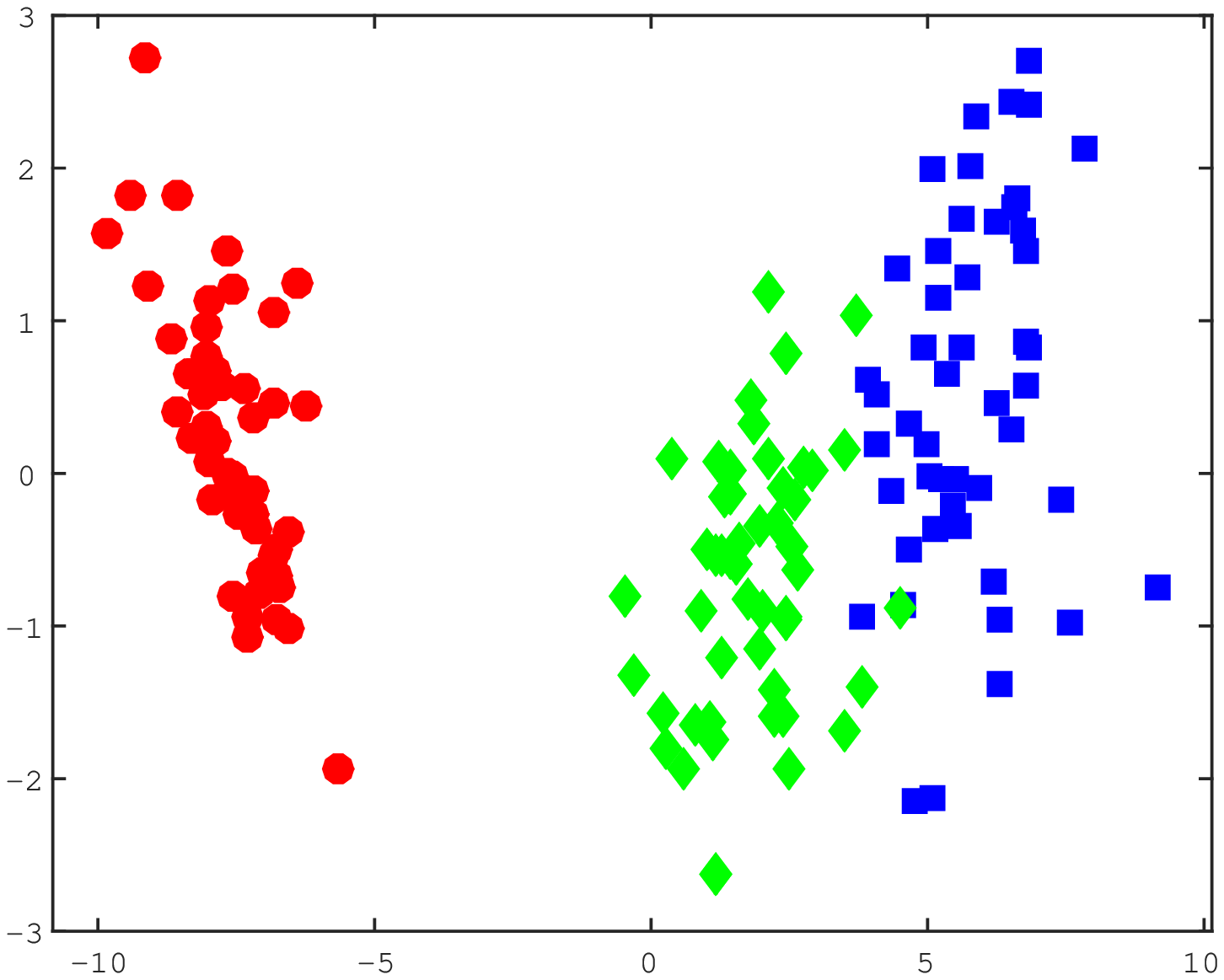}
       		 \caption{LDA (Iris): 2-D \label{fig:ldairis}}
 	 \end{center}
    \end{minipage}
\end{figure}

\begin{figure}[htbp]
  \begin{minipage}{0.5\hsize}
  	\begin{center}
		\includegraphics[width=6.5cm,height=5cm]{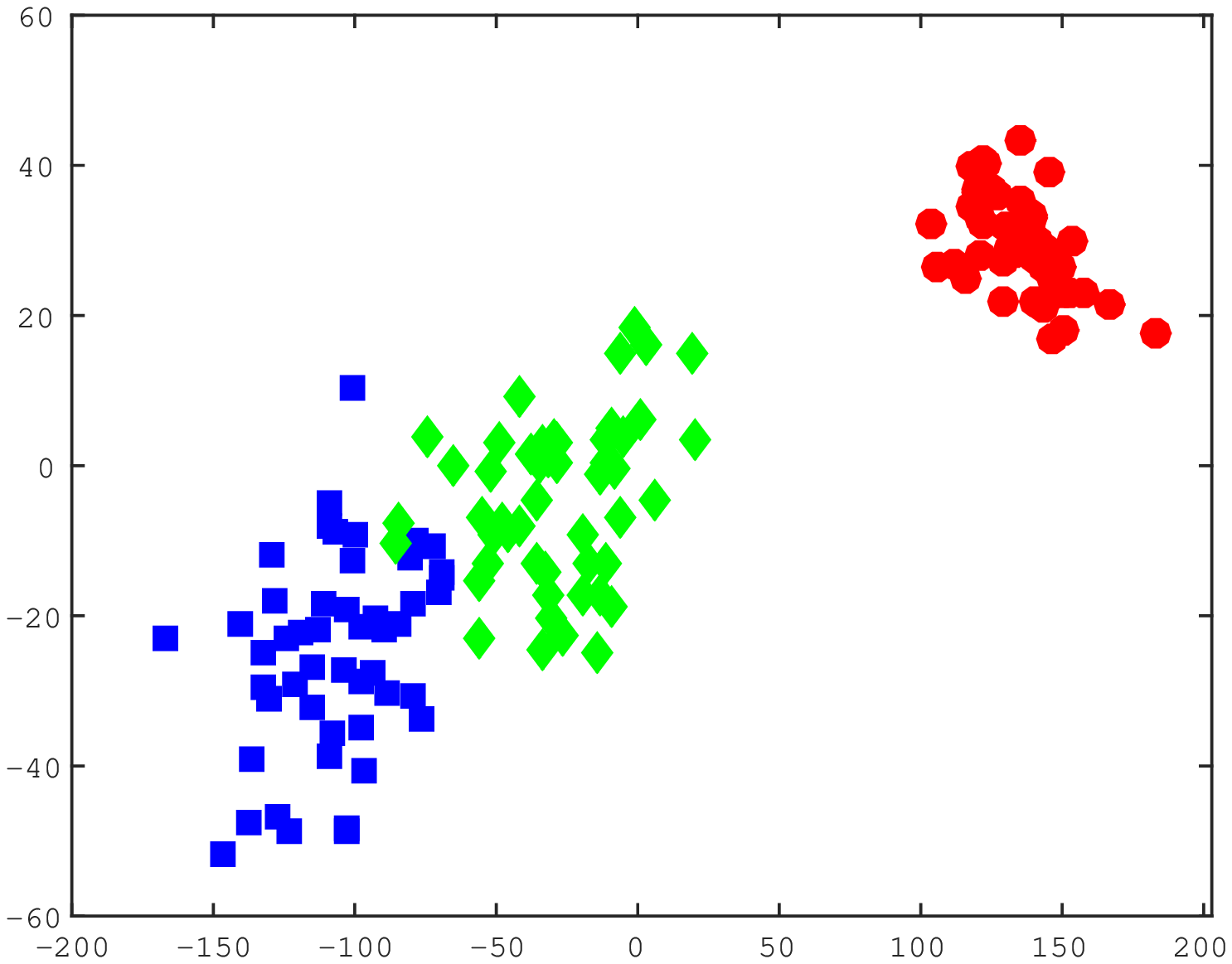}
       		 \caption{NCA (Iris): 2-D\label{fig:ncairis}}
 	 \end{center}
    \end{minipage}
   \begin{minipage}{0.5\hsize}
  	\begin{center}
		\includegraphics[width=6.5cm,height=5cm]{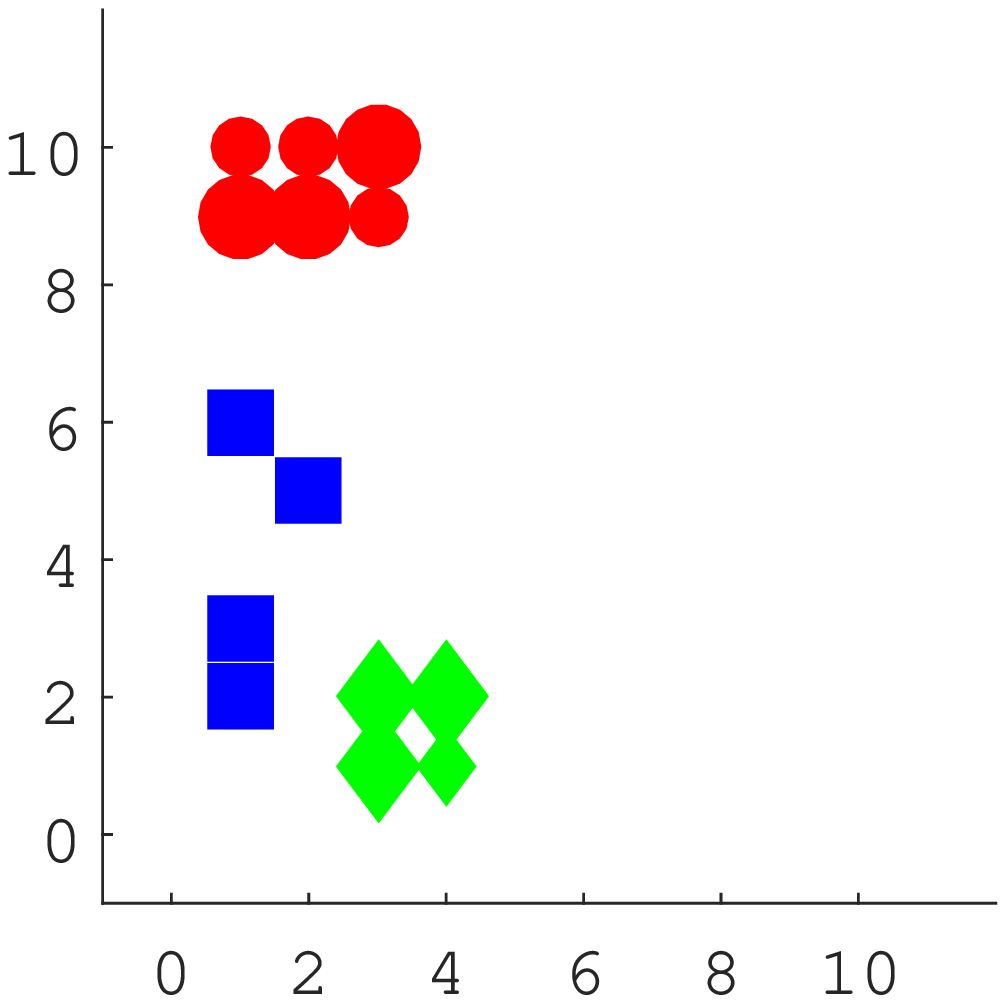}
       		 \caption{CRSOM (Iris)\label{fig:crsomiris}}
 	 \end{center}
    \end{minipage}
\end{figure}

Figures \ref{fig:pcairis}, \ref{fig:ldairis} and \ref{fig:ncairis} show the two dimensional projections of Iris Problem \cite{Fisher} through PCA, LDA and NCA respectively. This is a well known problem in which one of the class is linearly separable from the rests, while the other two are not. This class-characteristics are nicely captured in all the methods which are also indicated by the similar generalization performances, although rRBF performed slightly better.

\begin{figure}[htbp]
  \begin{minipage}{0.5\hsize}
  	\begin{center}
		\includegraphics[width=6.5cm,height=5cm]{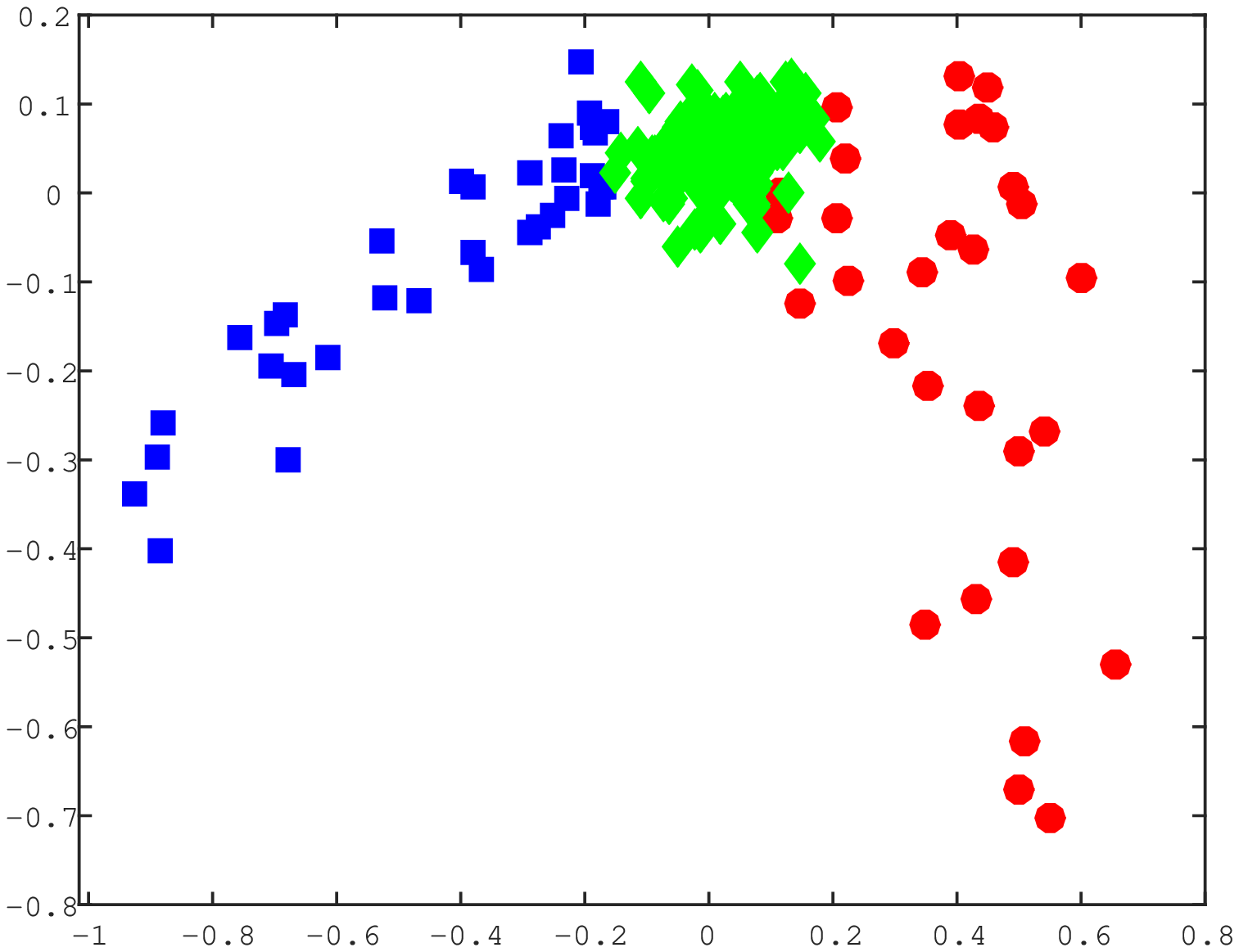}
       		 \caption{PCA (Thyroid): 2-D\label{fig:pcathyroid}}
 	 \end{center}
    \end{minipage}
   \begin{minipage}{0.5\hsize}
  	\begin{center}
		\includegraphics[width=6.5cm,height=5cm]{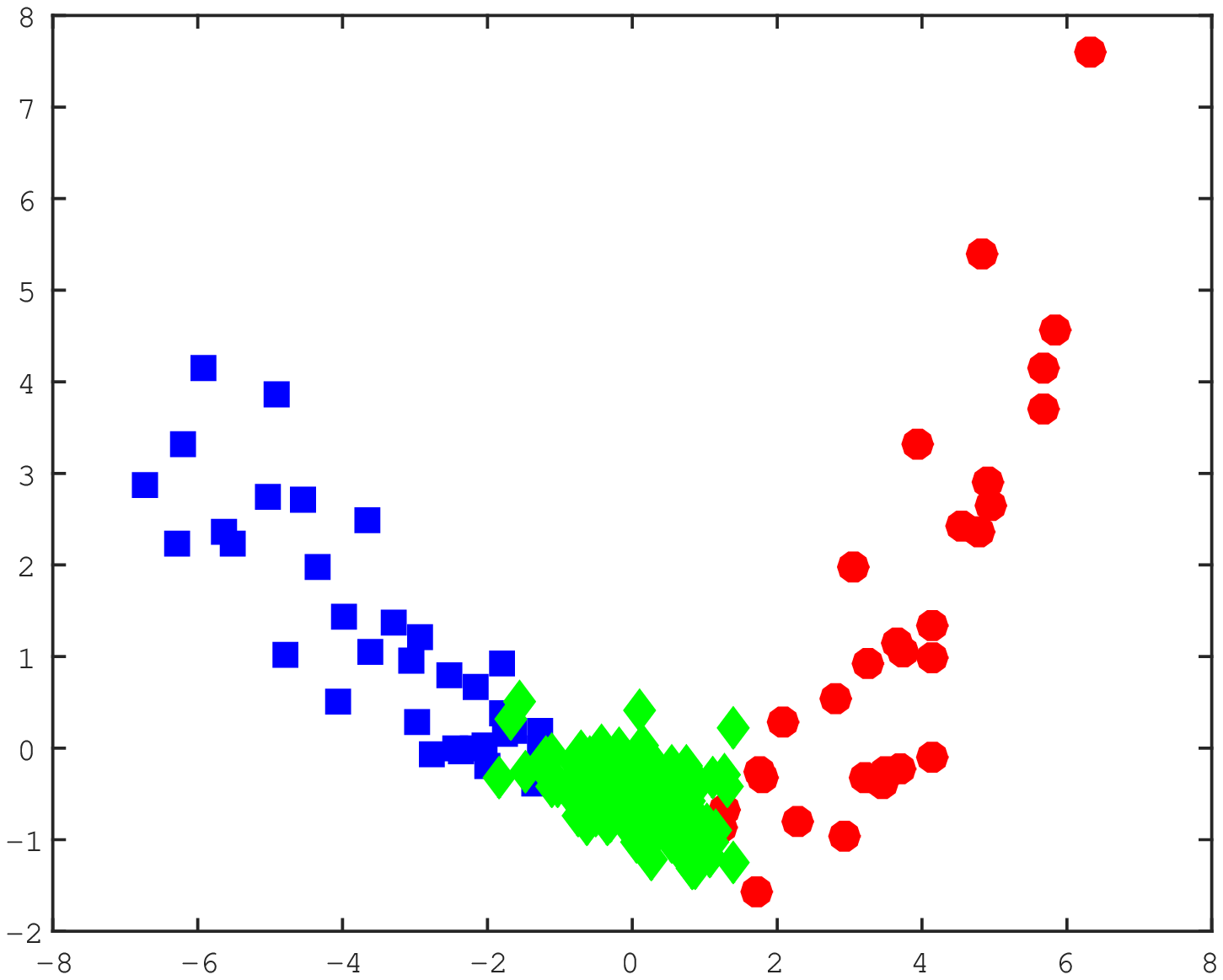}
       		 \caption{LDA (Thyroid): 2-D\label{fig:ldathyroid}}
 	 \end{center}
    \end{minipage}
\end{figure}

\begin{figure}[htbp]
  \begin{minipage}{0.5\hsize}
  	\begin{center}
		\includegraphics[width=6.5cm,height=5cm]{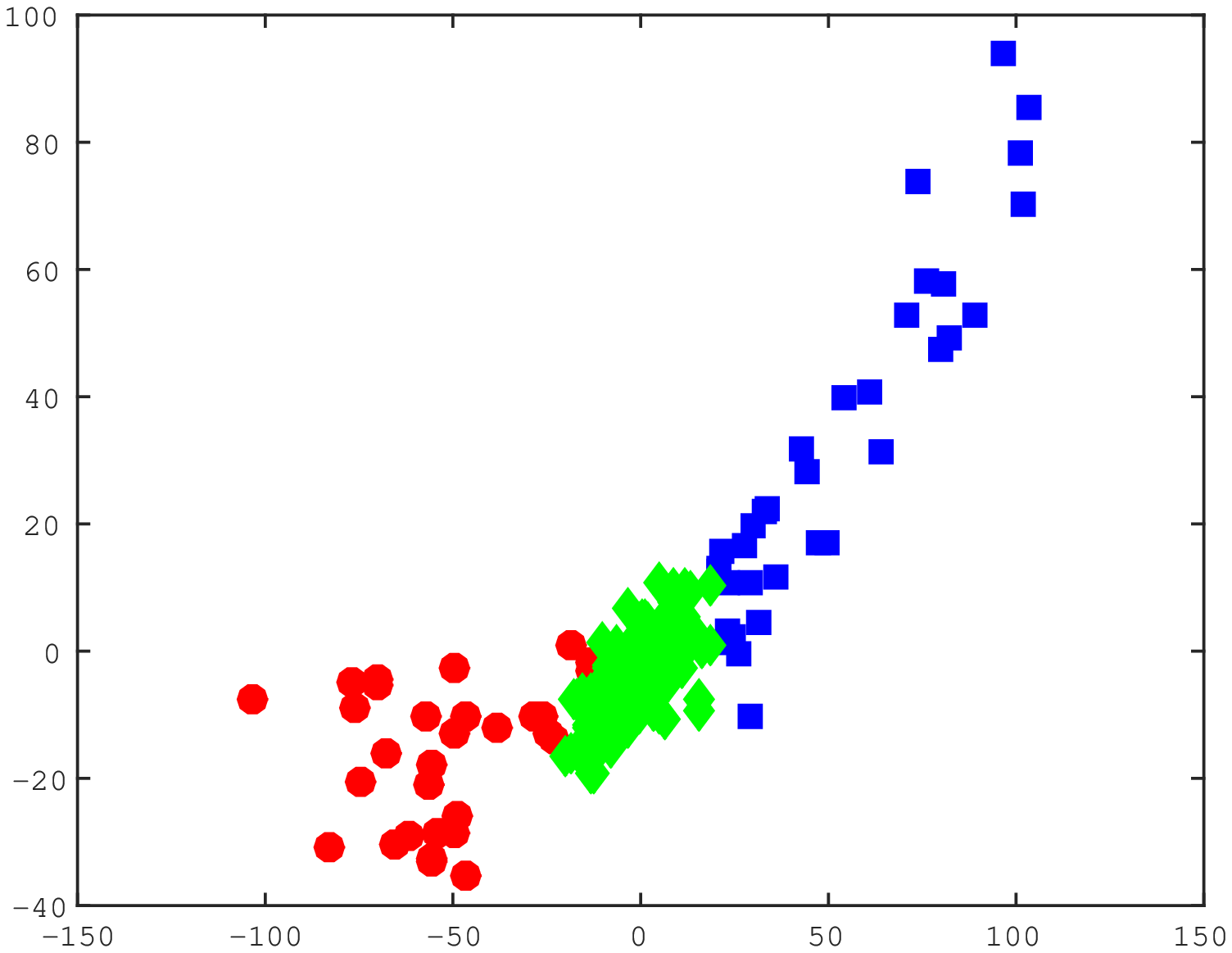}
       		 \caption{NCA (Thyroid): 2-D\label{fig:ncathyroid}}
 	 \end{center}
    \end{minipage}
   \begin{minipage}{0.5\hsize}
  	\begin{center}
		\includegraphics[width=6.5cm,height=5cm]{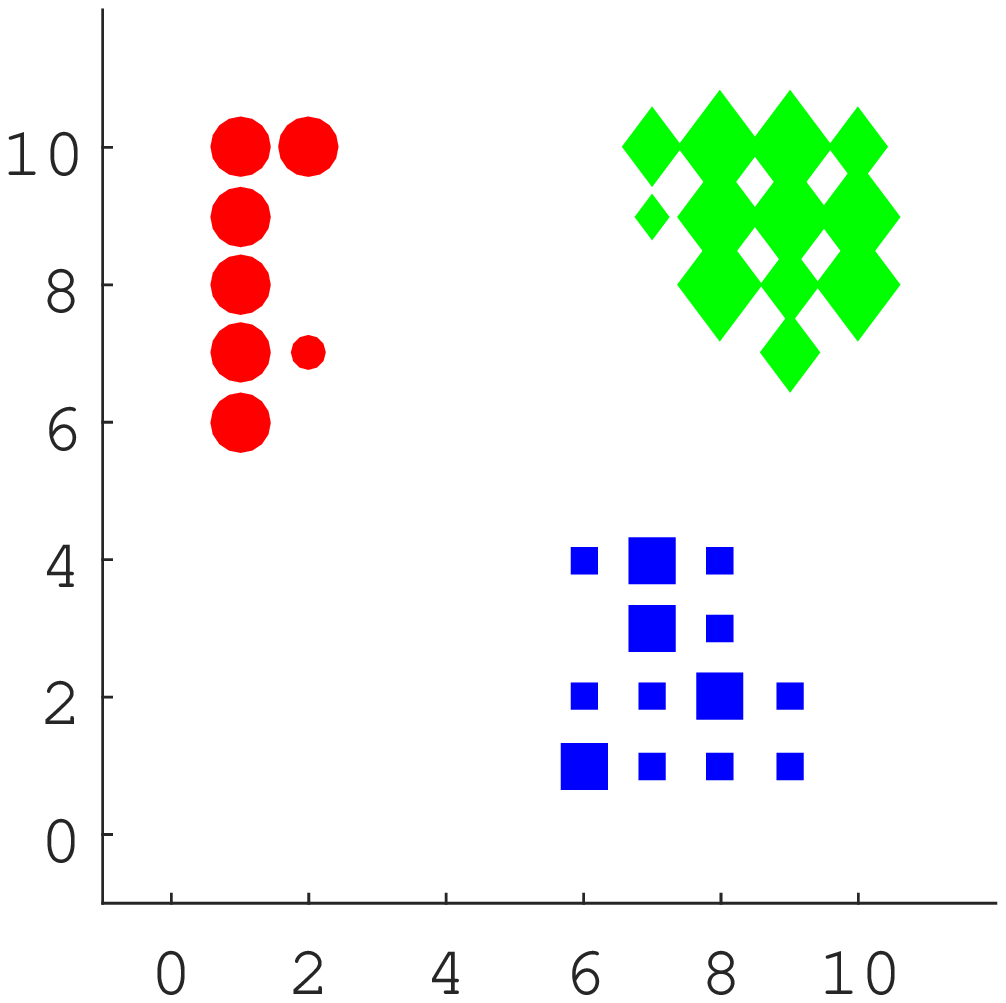}
       		 \caption{CRSOM (Thyroid)\label{fig:crsomthyroid}}
 	 \end{center}
    \end{minipage}
\end{figure}

Figures \ref{fig:pcathyroid}, \ref{fig:ldathyroid}, \ref{fig:ncathyroid} and \ref{fig:crsomthyroid} show the two dimensional projections of Thyroid Problem. This is an easy classification problem where each of the 3 classes forms a distinctive cluster, as illustrated by all of the compared methods. Consequently, all of the methods generate similar generalization performances, with two dimensional NCA performs slightly better.

\begin{figure}[htbp]
  \begin{minipage}{0.5\hsize}
  	\begin{center}
		\includegraphics[width=6.5cm,height=5cm]{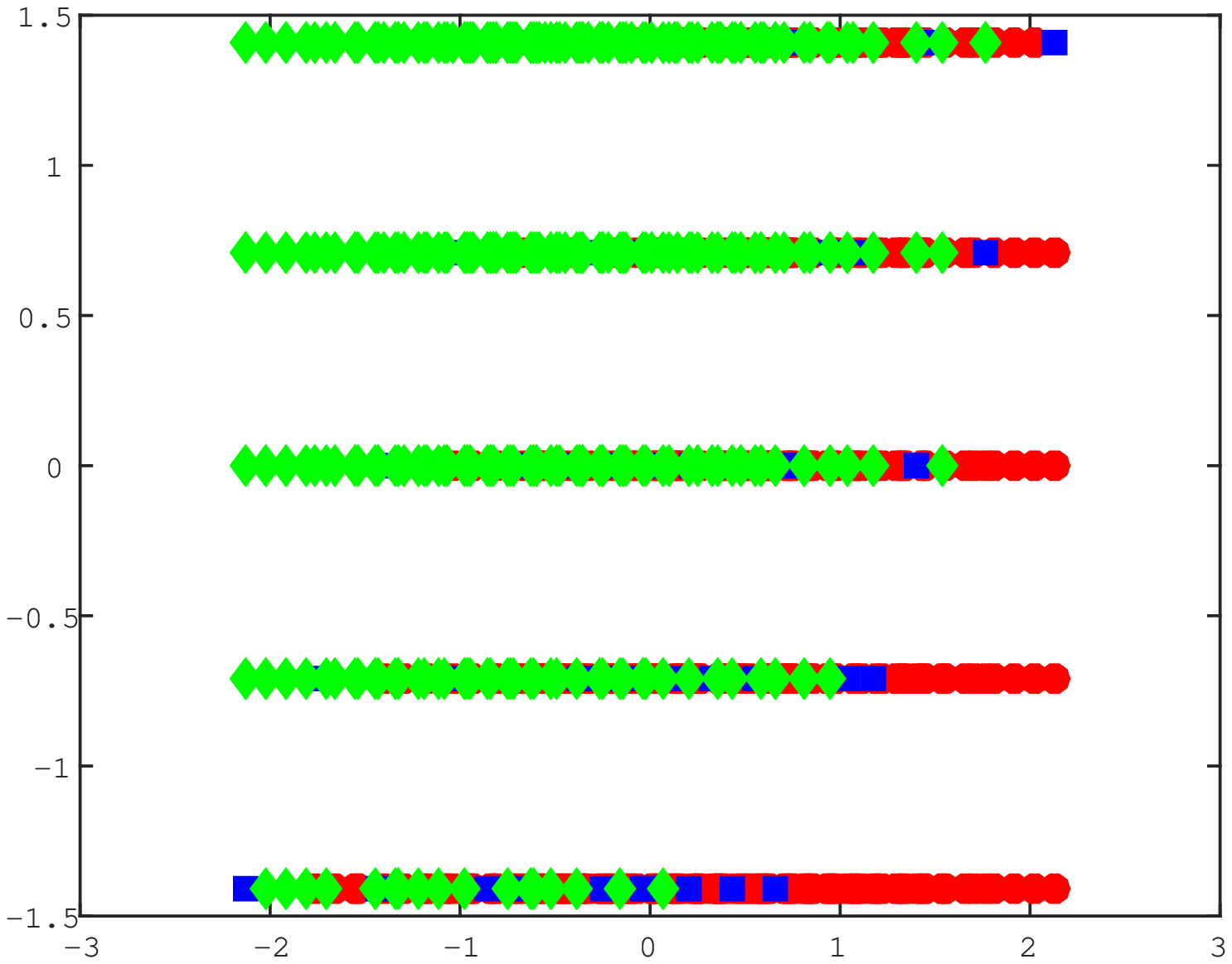}
       		 \caption{PCA (Balance): 2-D\label{fig:pcabalance}}
 	 \end{center}
    \end{minipage}
   \begin{minipage}{0.5\hsize}
  	\begin{center}
		\includegraphics[width=6.5cm,height=5cm]{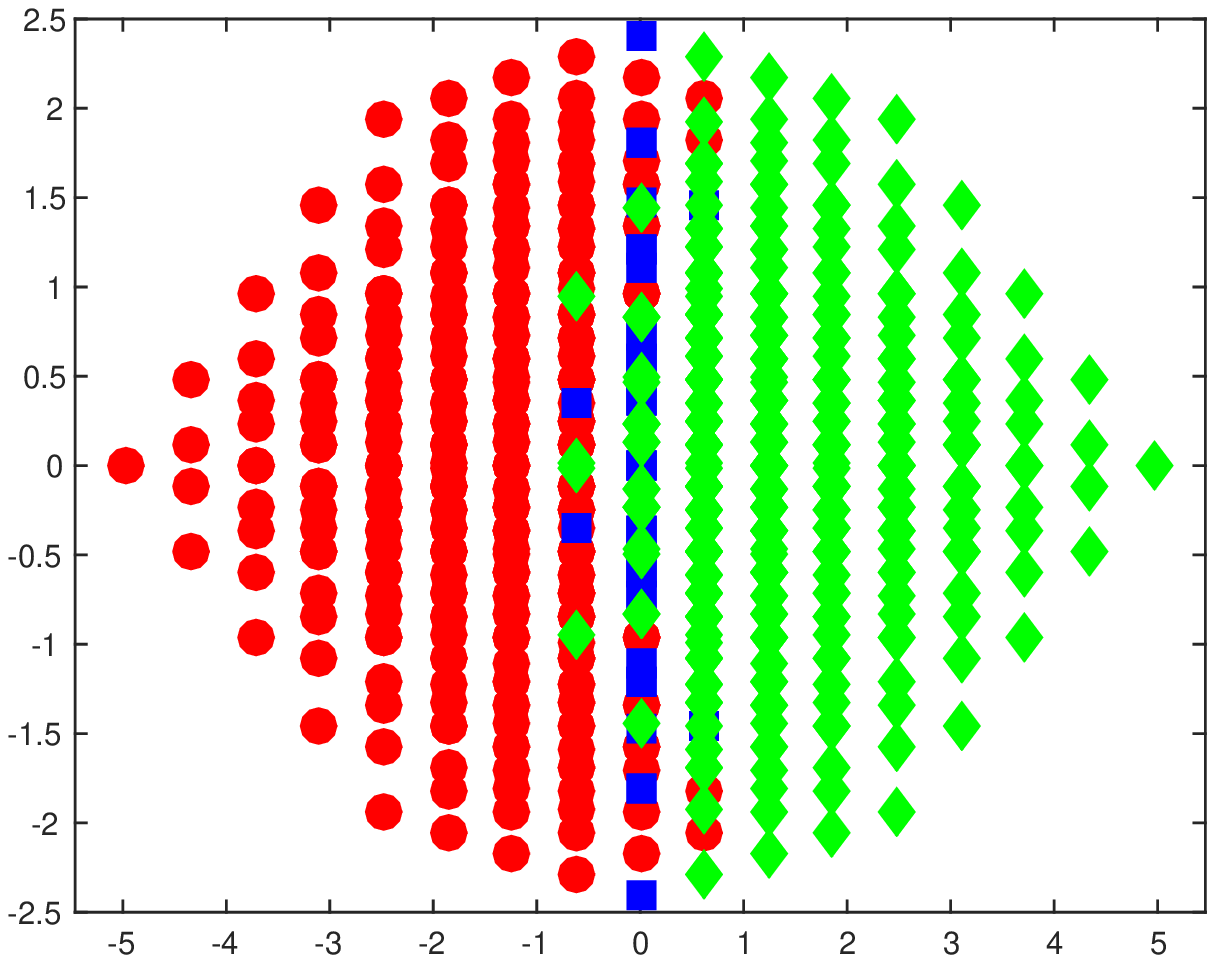}
       		 \caption{LDA (Balance): 2-D\label{fig:ldabalance}}
 	 \end{center}
    \end{minipage}
\end{figure}

\begin{figure}[htbp]
  \begin{minipage}{0.5\hsize}
  	\begin{center}
		\includegraphics[width=6.5cm,height=5cm]{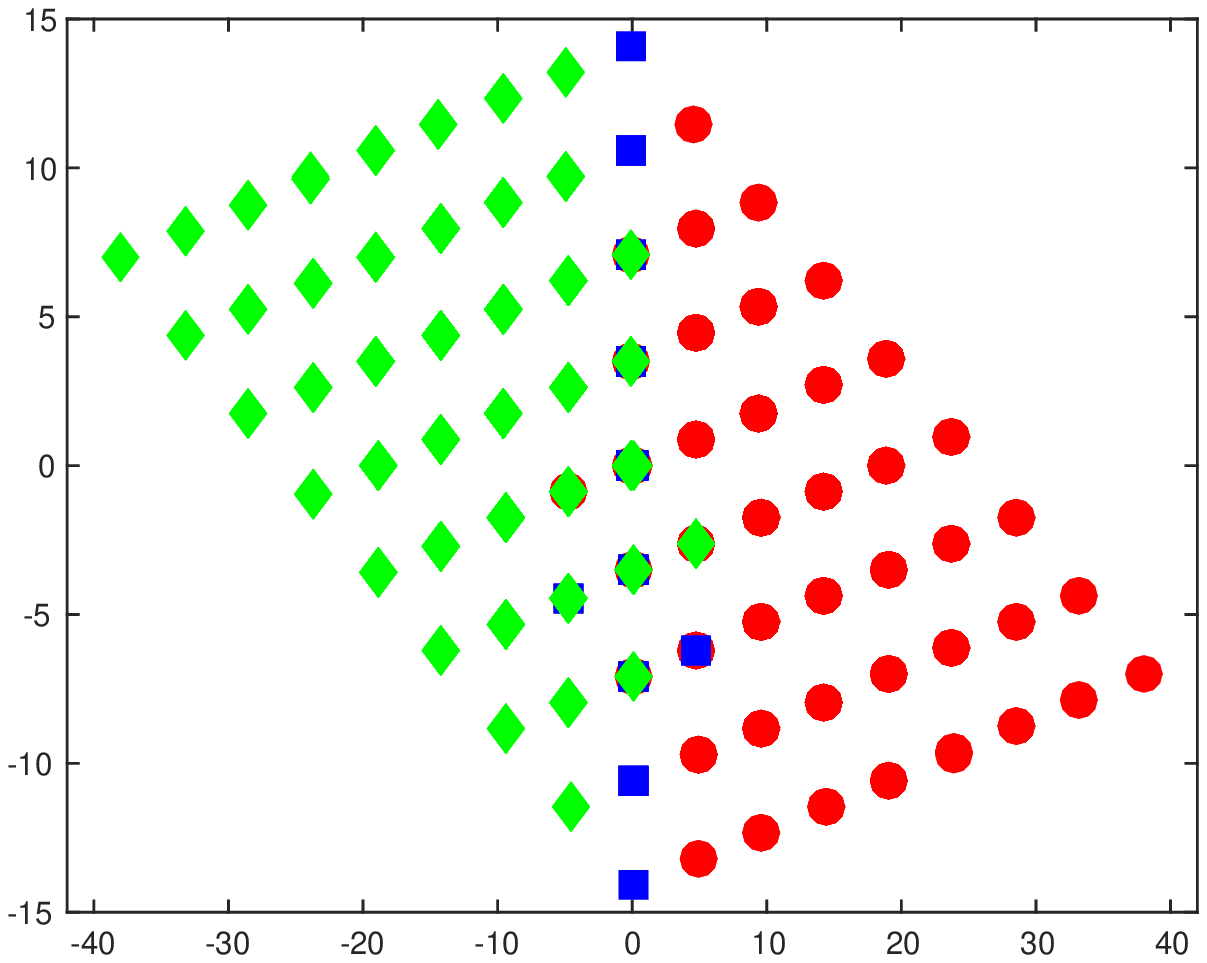}
       		 \caption{NCA (Balance): 2-D\label{fig:ncabalance}}
 	 \end{center}
    \end{minipage}
  \begin{minipage}{0.5\hsize}
  	\begin{center}
		\includegraphics[width=6.5cm,height=5cm]{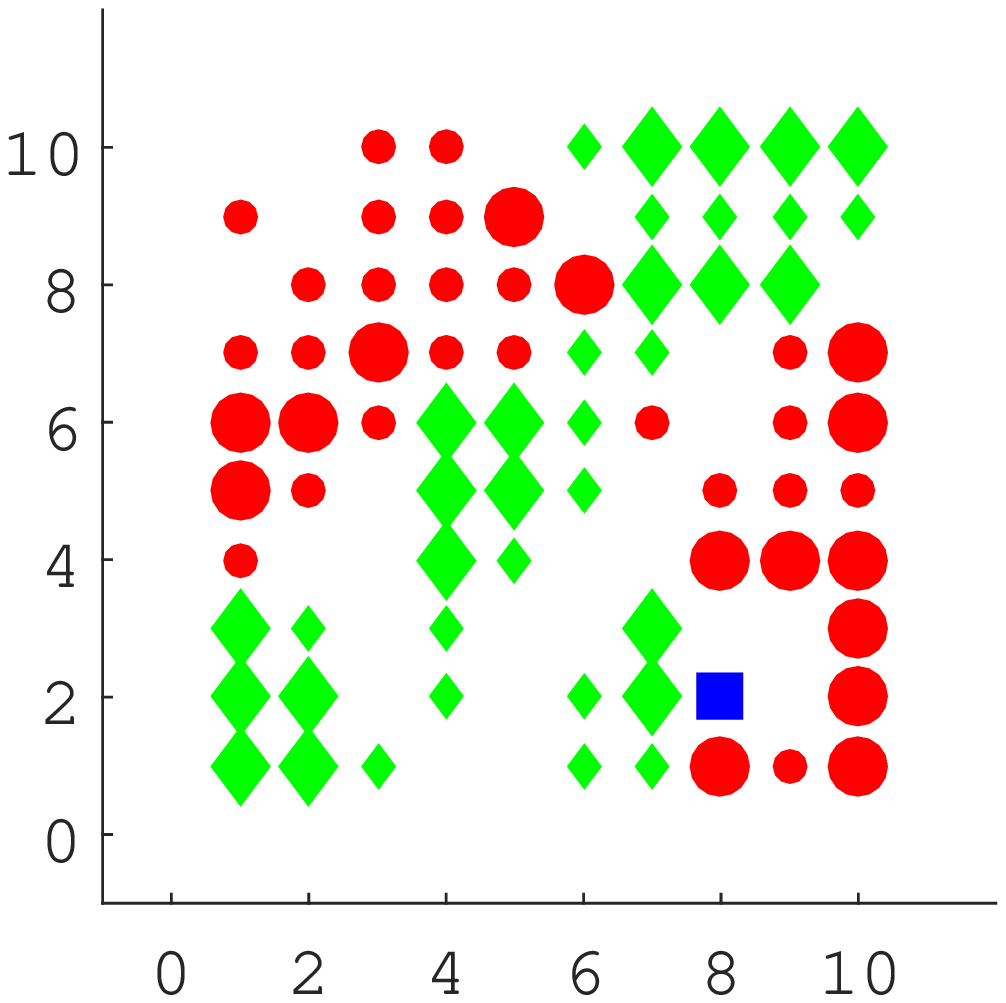}
       		 \caption{CRSOM (Balance) \label{fig:crsombalance}}
 	 \end{center}
    \end{minipage}
\end{figure}

Figures \ref{fig:pcabalance}, \ref{fig:ldabalance}, \ref{fig:ncabalance} and \ref{fig:crsombalance} show the two dimensional representations of Balance problem. This problem is interesting in that all the compared methods generated visually different representations. Table \ref{tab:error} indicates that the rRBF performed worst than the best performing NCA but still has significantly better result than two dimensional PCA. In this problem, one of the three classes is represented by excessively low number of samples, which is clearly captured by the CRSOM, where the underrepresented class is depicted by only one winning neuron that is surrounded by neurons belonging to the other classes. The CRSOM in Fig. \ref{fig:crsombalance} is interesting in that it represents one of the classes with two separate clusters, a class neighborhood property that is not captured by other dimension reduction methods.

\begin{figure}[htbp]
  \begin{minipage}{0.5\hsize}
  	\begin{center}
		\includegraphics[width=6.5cm,height=5cm]{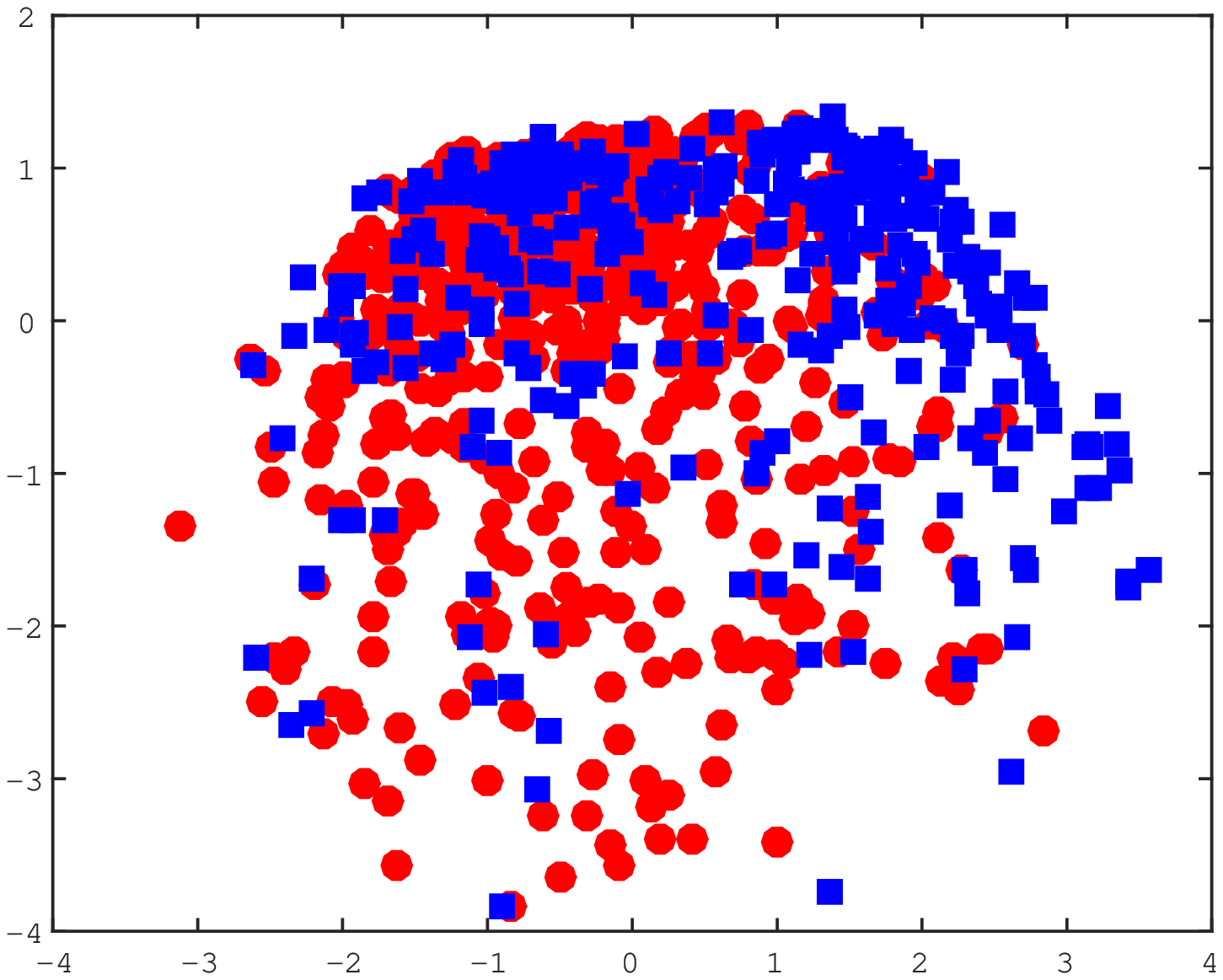}
       		 \caption{PCA (music): 2-D\label{fig:pcamusic}}
 	 \end{center}
    \end{minipage}
   \begin{minipage}{0.5\hsize}
  	\begin{center}
	 	 \includegraphics[width=6.5cm,height=5cm]{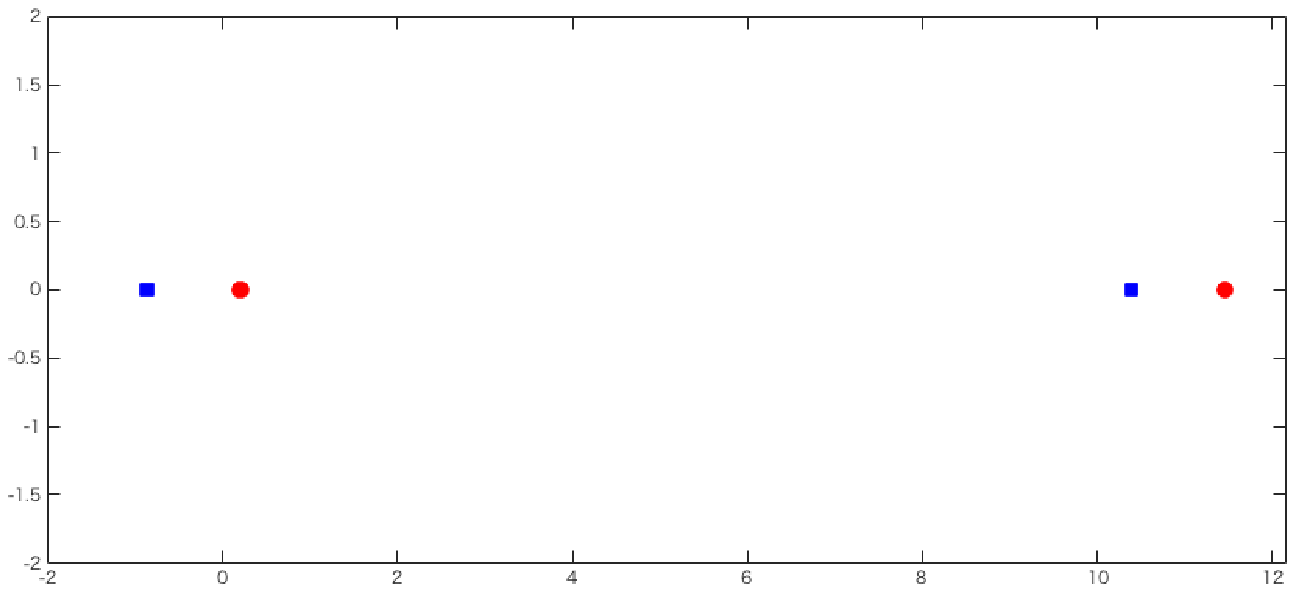}
       		 \caption{LDA (music): 2-D\label{fig:ldamusic}}
 	 \end{center}
    \end{minipage}
\end{figure}

\begin{figure}[htbp]
  \begin{minipage}{0.5\hsize}
  	\begin{center}
		\includegraphics[width=6.5cm,height=5cm]{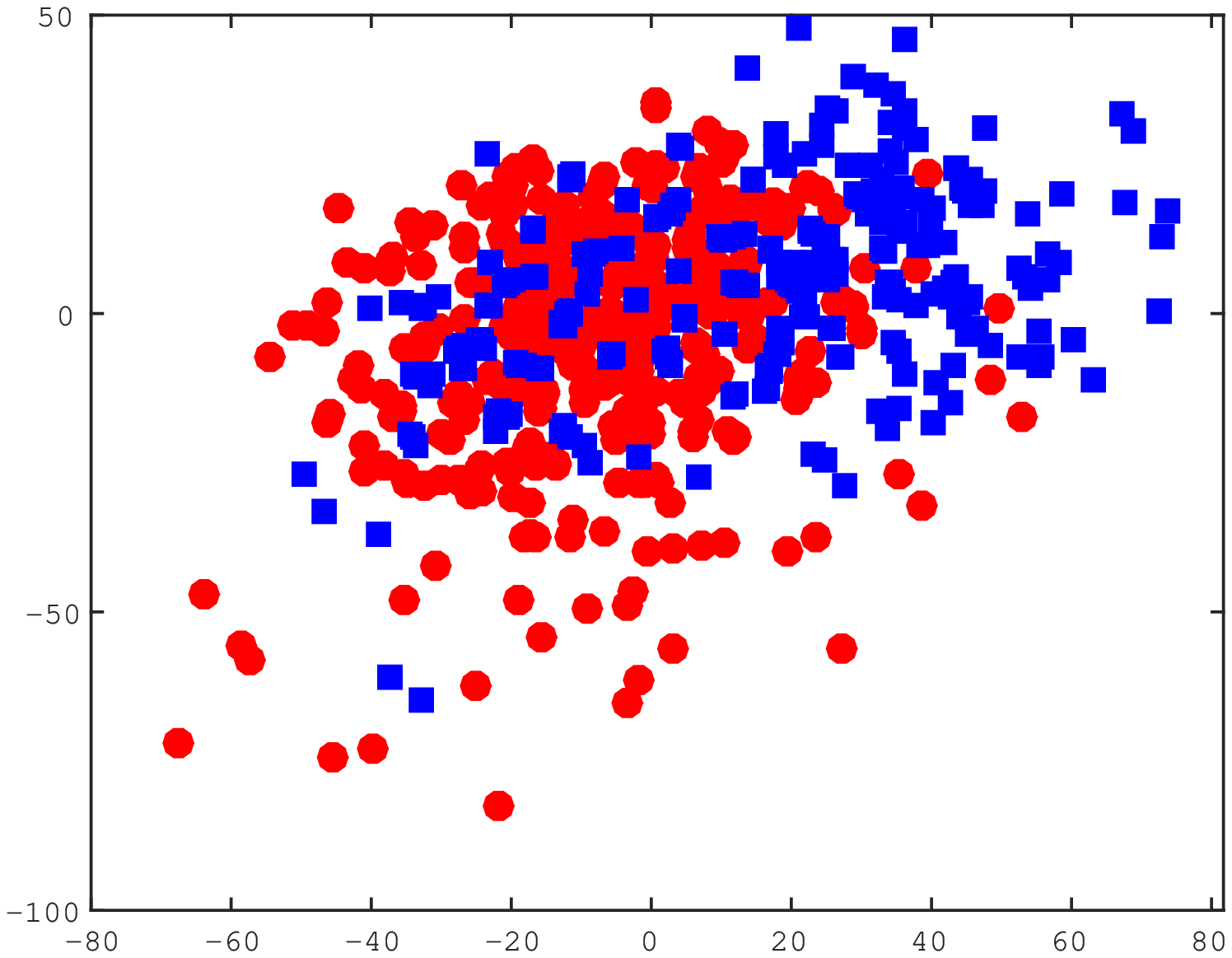}
       		 \caption{NCA (music): 2-D\label{fig:ncamusic}}
 	 \end{center}
    \end{minipage}
  \begin{minipage}{0.5\hsize}
  	\begin{center}
		\includegraphics[width=6.5cm,height=5cm]{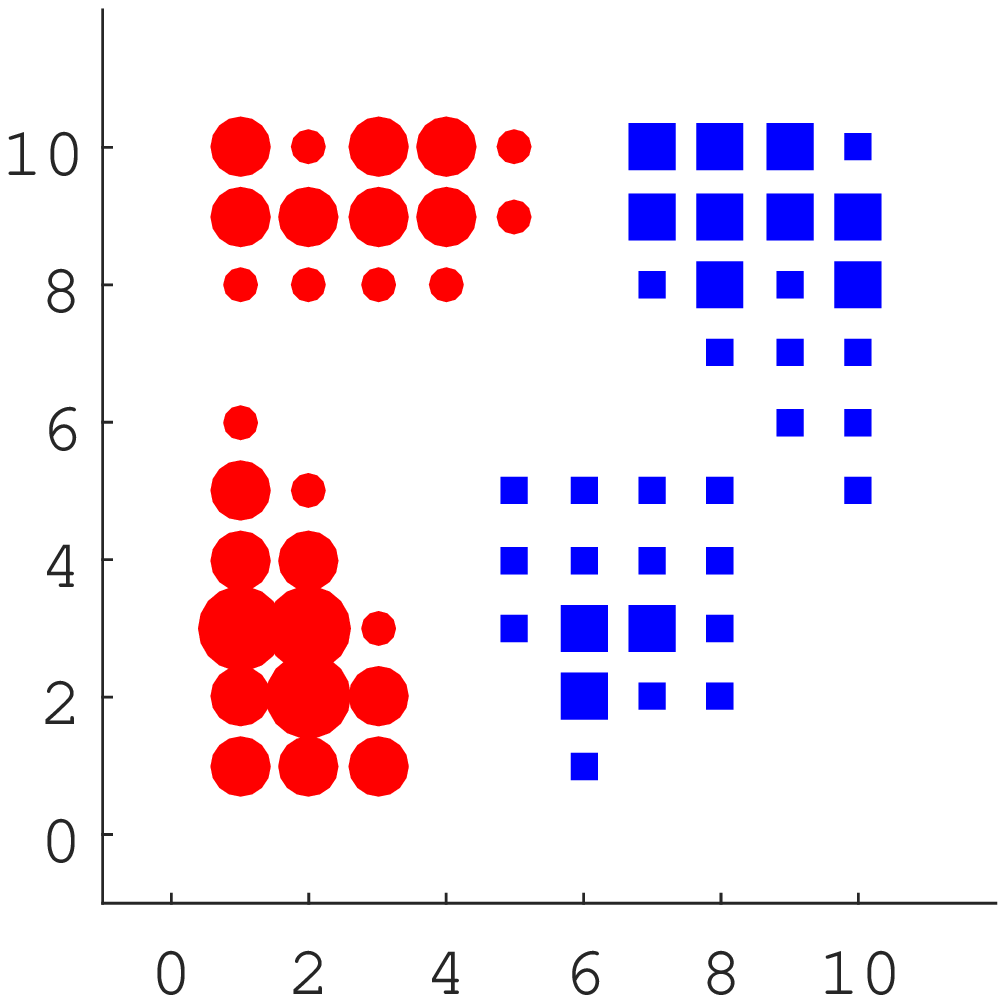}
       		 \caption{CRSOM (music) \label{fig:crsommusic}}
 	 \end{center}
    \end{minipage}
\end{figure}

Figures \ref{fig:pcamusic} and \ref{fig:ncamusic}, are the two dimensional projection of 2508 dimensions, 2 classes Music data used for psychological bi-musical experiment on PCA and NCA-reduced space, while Fig. \ref{fig:ldamusic} is the LDA's two dimensional projection of this problem. As this is a two classes classification problem, LDA transfers the data into a line, where the data are projected into two opposite ends of the line, in which subsets from both classes are aligned very close to each other, resulting in high error rate as indicated in Table \ref{tab:error}.

The CRSOM generates a representation where each classes is distinctively represented by two two separate clusters which clearly results in significantly high generalization performances compared to other two dimensional representations that failed to untangle the two different classes. 

\begin{figure}[htbp]
  \begin{minipage}{0.5\hsize}
  	\begin{center}
		\includegraphics[width=6.5cm,height=5cm]{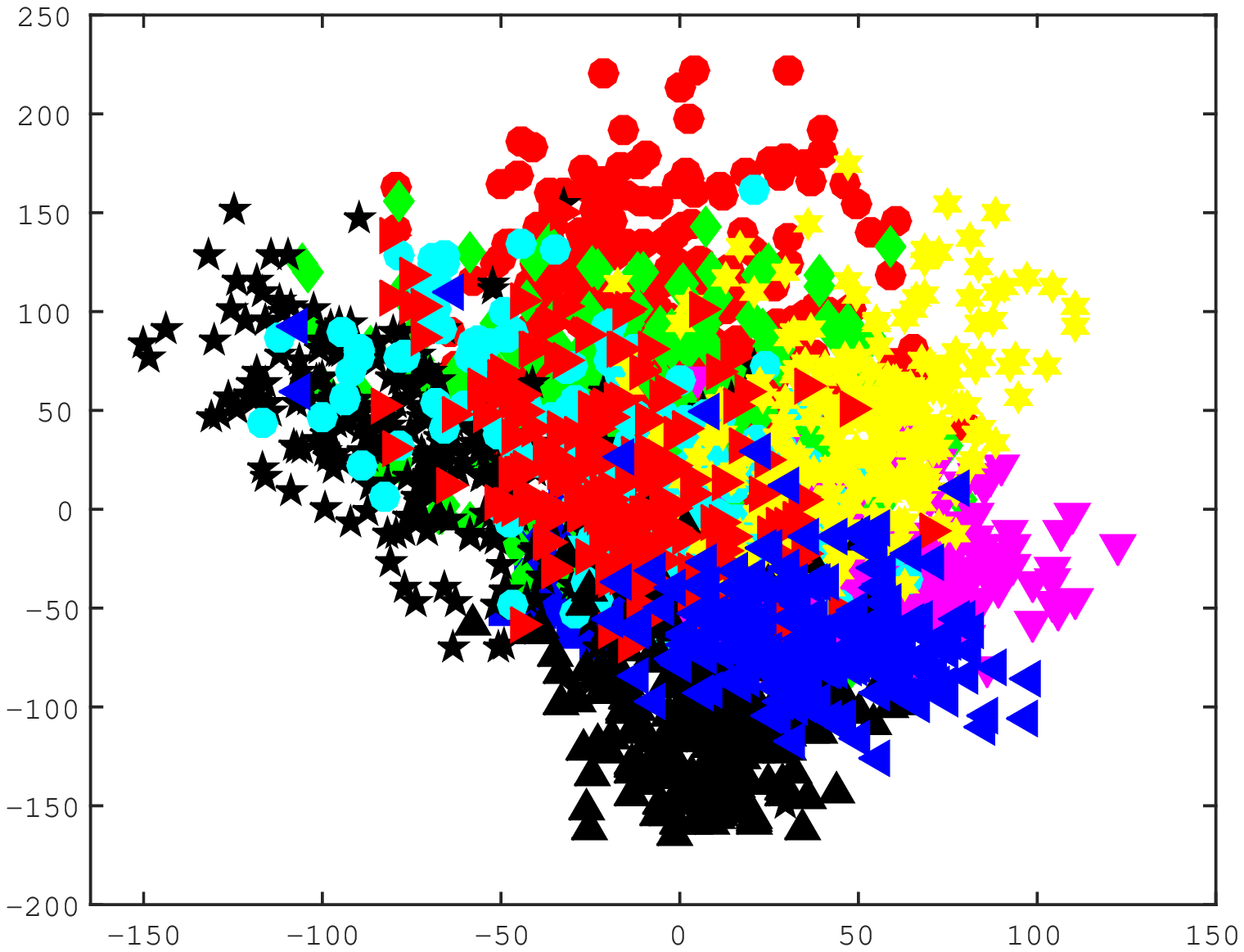}
       		 \caption{NCA (MNIST): 2-D\label{fig:ncamnist}}
 	 \end{center}
    \end{minipage}
   \begin{minipage}{0.5\hsize}
  	\begin{center}
		\includegraphics[width=6.5cm,height=5cm]{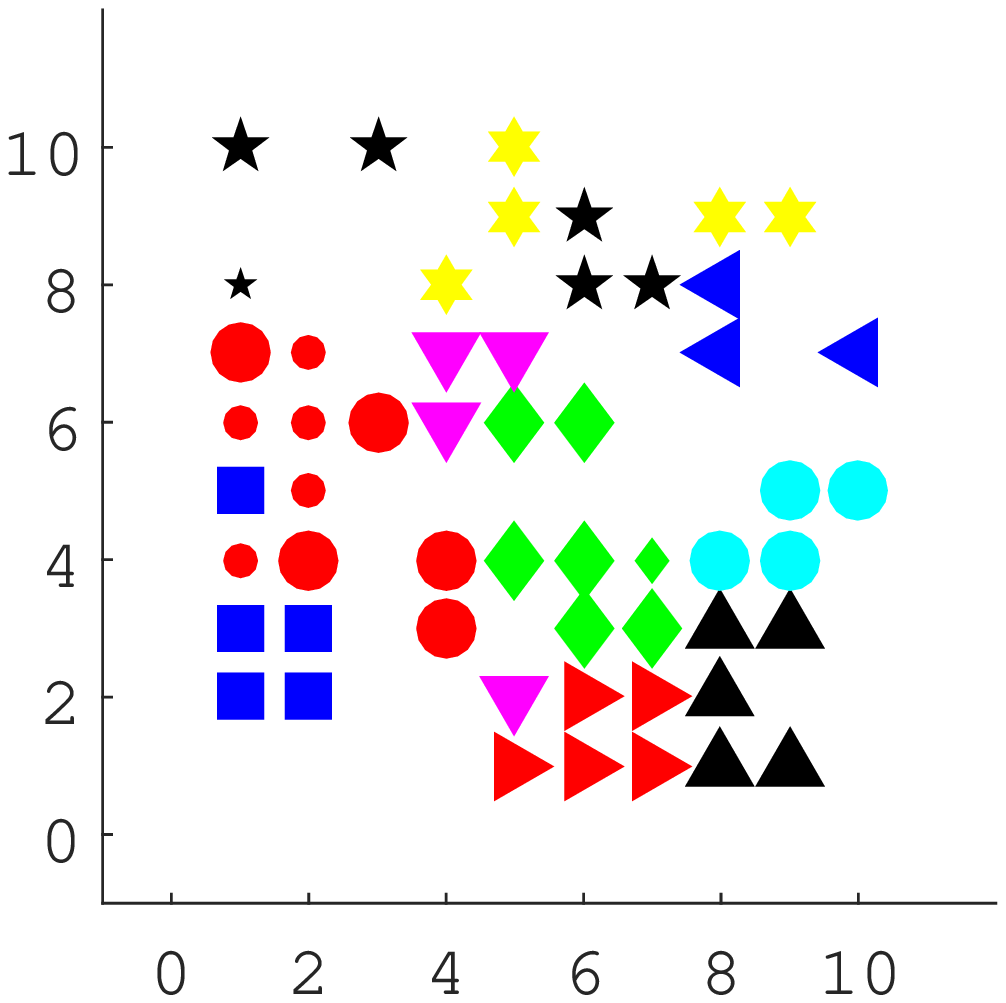}
       		 \caption{CRSOM (MNIST\label{fig:crsommnist}}
 	 \end{center}
    \end{minipage}
\end{figure}

\begin{figure}[htbp]
  \begin{center}
      \includegraphics[width=10cm,height=5cm]{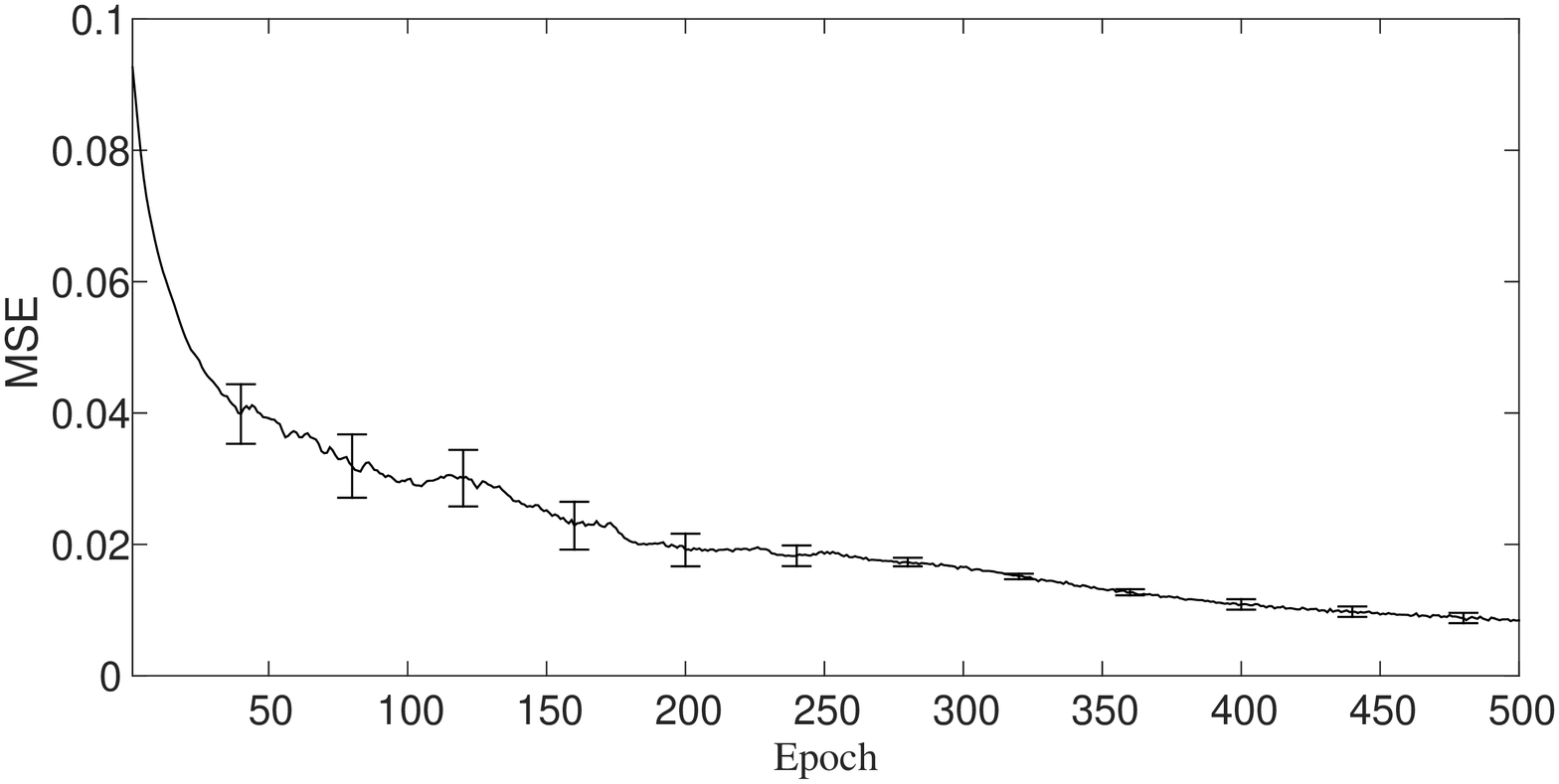}
      \caption{ Learning Curve (MNIST) \label{fig:learnmnist}}
  \end{center}
\end{figure}

The NCA projection and CRSOM for MNIST problem are shown in Fig. \ref{fig:ncamnist} and Fig. \ref{fig:crsommnist}, while the PCA's and LDA's two dimensional representations are shown in Figures \ref{fig:pcamnist} and \ref{fig:ldamnist}. It is obvious that the rRBF generated a two dimensional representation where the classes are distinctively separated, while other two dimensional representations failed to do so. In two dimensional space, rRBF performed significantly better than the other methods. From Table \ref{tab:error} it can be learned that the generalization performance of the rRBF is very close to the generalization performed in the original dimensional space. This performance similarity is an indication that CRSOM has the ability to preserve the underlying class neighborhood properties of high dimensional data in visualizable low dimensional space while other dimensional reduction methods often fail to do so. As the organization of the low dimensional representation of the training data for  the rRBF is based on error minimization, Fig. \ref{fig:learnmnist} is presented to depict the average of the learning process over 10 different runs.

In the experiments, the number of the nearest neighbors for classifications was uniformly and empirically set to 3 because in average it produced the best classification rates over all the problems. The classification rates for each problem slightly differs with the change of the number of the nearest neighbors, but not significantly. For the same reason, the size of the CRSOM for all of the problems was set to $10 \times 10$, while the learning iterations for the rRBF was set to 500. $S_{start}$ and $S_{end}$ in Eq. \ref{eq:neighbor} are set to 50 and 0.01, respectively.

\section{Conclusions}
In this paper, an empirical analysis on the non-degrading generalization of dimension reduction of the rRBF is explained. The experiments indicate that the nearest neighbors classification of categorical data on the reduced-dimension often gives poor results. The primary cause for the poor results is that many of the attributes are often uncorrelated, hence reducing them into visualizable dimensional size causes the lost of many important features for correctly classifying the data. In the experiment, this property was shown by the significant degradations of the generalization performances along with the decreasing dimension. For visualizing high dimension data, the dimension of the data has to be reduced, but in this case, the low dimensional appearance of the data offers little insight as the class neighborhood structure is not well represented.

The rRBF offers visualization in two-dimensional space by preserving not only the topographical structure of the data but also their class neighborhood structure, thus it visualizes not only the data but the problem. The mathematical derivation of the learning process shows that the CRSOM is an optimal representation of the high dimensional data in the two dimensional internal layer of the rRBF. This infers that the appearance of the CRSOM is directly related with the classification process of the rRBF. The generalization performance of the rRBF may be inferior to deeper layer networks, but its structural simplicity is an advantage in executing faster leaning and it also presents mathematically more comprehensive two dimensional representation of high dimensional data, in that the derivation of the learning process clearly indicate the formation of context-relevant topological structure. 

As the CRSOM generates sparse representations, the future works include the investigation of the learning properties of rRBF in avoiding the catastrophic forgetting that is known to occur in hierarchical neural networks with non-sparse representation.The ability of the rRBF in executing incremental learning will also be thoroughly studied, while the investigation on the different formations of the  internal representations with regards to different learning algorithm, for example reinforcement learning, will also be considered in the future.

\section*{Acknowledgment}
The author would like to thank Dr. Rie Matsunaga from Shizuoka Institute of Science and Technology for providing the psychological experiment data for Music dataset.


\bibliography{mybibfile}

\end{document}